\documentclass[preprint,authoryear,12pt]{elsarticle}
\usepackage[utf8]{inputenc}
\usepackage{amsmath}
\usepackage{amsfonts}
\usepackage{amssymb}
\usepackage{graphicx}
\usepackage{ifthen}
\usepackage{url}
\usepackage{hyperref}

\newcommand{\Problem}[1]{\textsc{#1}}
\newcommand{\PProblem}[2]{$\langle$\textsc{#1},$#2$$\rangle$}

\def\HD{\ensuremath{{\cal{HD}}}}

\newcommand{\subindex}[2]{\ensuremath{{#1}_{\mbox{\tiny{#2}}}}}

\newcommand{\refRange}[3]{{#1}~\ref{#2}--\ref{#3}}
\newcommand{\refAndPage}[2]{{#2}~\ref{#1}}
\newcommand{\refTable}[1]{\refAndPage{#1}{Table}}
\newcommand{\refFig}[1]{\refAndPage{#1}{Figure}}
\newcommand{\refFigs}[2]{\refRange{Figures}{#1}{#2}}

\newcommand{\refEq}[1]{\refAndPage{#1}{Equation}}

\newcommand{\kw}[1]{\textbf{#1}}
\newcommand{\fnctn}[1]{\textsc{#1}}

\newcounter{AlgCounter}
\newcommand{\LnNumber}{\\[0.3mm] \> \> \small{\theAlgCounter}\,:\' \> \stepcounter{AlgCounter}}
\newcommand{\LnFunction}{\\[0.3mm] \> \>}
\newcommand{\LnNoNumber}{\\[0.3mm] \> \> \>}
\def\oneTab{\ \ \ \=}
\newenvironment{algorithm}[3]%
{\setcounter{AlgCounter}{1}%
\centering%
\rule{0.97\textwidth}{0.5pt}%
\vspace*{-0.7\baselineskip}%
\begin{center}%
\textbf{#1}\\[-3mm]%
\rule{0.97\textwidth}{0.5pt}%
\end{center}%
\vspace*{-13mm}%
\begin{tabbing}%
\ \
\oneTab\oneTab\oneTab\oneTab\oneTab\oneTab\oneTab\oneTab\oneTab\oneTab\oneTab\oneTab\oneTab\kill
\ifthenelse{\not{\equal{}{#2}}}{%
\\%
\> \textsc{Input:}  \> \> \> \> \> \> {#2}.\\%
\> \textsc{Output:} \> \> \> \> \> \> {#3}.%
\\[-0.75\baselineskip]
}% else
{}%
}%
{\end{tabbing}%
\vspace{-2.0\baselineskip}%
\begin{center}\rule{0.97\textwidth}{0.75pt}\end{center}%
\vspace{-\baselineskip}%
}

\def\GRASPFFR{GRASP$_{\mathrm{FFR}}$}
\def\GRASPMou{GRASP$_{\mathrm{Mou}}$}
\def\MA{MA$_{\mathrm{GRASP+HC+PR}}$}

\newcommand{\ma}[3]{MA$_{\mathrm{GRASP_{#1}+{#2}{#3}}}$}

\newcommand{\maPRHC}[1]{\ma{#1}{PR}{+HC}}
\newcommand{\maUXHC}[1]{\ma{#1}{UX}{+HC}}
\newcommand{\maPR}[1]{\ma{#1}{PR}{}}
\newcommand{\maUX}[1]{\ma{#1}{UX}{}}

\newcommand{\maPRHCRND}{MA$_{\mathrm{RND+PR+HC}}$}
\newcommand{\maUXHCRND}{MA$_{\mathrm{RND+UX+HC}}$}
\newcommand{\maPRRND}{MA$_{\mathrm{RND+PR}}$}
\newcommand{\maUXRND}{MA$_{\mathrm{RND+UX}}$}

\begin{document}

\begin{frontmatter}

\title{A GRASP-based Memetic Algorithm with Path Relinking for the Far From Most String Problem}

\author{Jos\'{e}~E.~Gallardo\corref{cor1}}
\ead{pepeg@lcc.uma.es}
\author{Carlos Cotta}
\ead{ccottap@lcc.uma.es}

\cortext[cor1]{Corresponding author}

\address{
    Department {\em Lenguajes y Ciencias de la Computaci\'on},
    Universidad de M\'alaga,
    Campus de Teatinos, 29071 - M{\'a}laga, Spain.}

\begin{abstract} The \Problem{Far From Most String Problem} (FFMSP) is a string selection problem. The objective is to find a string whose distance to other strings in a certain input set is above a given threshold for as many of those strings as possible. This problem has links with some tasks in computational biology and its resolution has been shown to be very hard. We propose a memetic algorithm (MA) to tackle the FFMSP. This MA exploits a heuristic objective function for the problem and features initialization of the population via a Greedy Randomized Adaptive Search Procedure (GRASP) metaheuristic, intensive recombination via path relinking and local improvement via hill climbing. An extensive empirical evaluation using problem instances of both random and biological origin is done to assess parameter sensitivity and draw performance comparisons with other state-of-the-art techniques. The MA is shown to perform better than these latter techniques with statistical significance.
\end{abstract}

\begin{keyword}
    Far from most string problem
    \sep String selection problems
		\sep Bioinformatics
    \sep Metaheuristics
		\sep Memetic algorithms.
\end{keyword}

\end{frontmatter}

{\footnotesize
\noindent\url{http://dx.doi.org/10.1016/j.engappai.2015.01.020}

\noindent © 2014. This manuscript version is made available under the CC-BY-NC-ND 4.0 license\\ \url{https://creativecommons.org/licenses/by-nc-nd/4.0}
}

\section{Introduction}
\label{sec:intro}

From a very general point of view, \emph{string selection problems} (SSPs)
can be defined as a class of problems involving the construction or
identification of a string of symbols satisfying certain constraints (typically distance constraints) with respect to another certain set of
strings provided as input. Such problems have attracted a lot of interest
for multiple reasons. From a theoretical (and even from a purely algorithmic)
 point of view, they constitute a clear and well-defined domain in which
computational complexity issues can be analyzed and search/optimization
algorithms can be put to work in challenging conditions. From a more
practical point of view, there are many real-world problems which can be
formalized as SSPs. Such problems are notably found in the area of
computational biology, in which technological advances and the numerous
initiatives are producing an unprecedented
flood of data \citep{Reichhardt_99} very much requiring the use of powerful
computational tools to overcome the associated challenges \citep{Meneses05optimization}.
Among such problems of interest from the perspective of SSPs
we can cite discovering potential drug targets, creating diagnostic probes,
designing primers, locating binding sites, or identifying consensus sequences
just to name a few \citep{Festa2007219,KevinLanctot200341,Meneses05optimization}.

SSPs can be classified in different ways. Following \cite{pappalardo13optimization},
we can consider median SSPs (in which the
goal is to find a string that minimizes the sum of distances to strings in the
input set), closest SSPs (aiming to identify a string
that is close to strings in the input set or to fragments thereof), and
farthest SSPs (whose goal is the recognition of
differences between sequences). Distinguishing SSPs have
also been defined \citep{KevinLanctot200341} and can be regarded as a
combination of closest and farthest SSPs in which the
input set is partitioned in two subsets, and a string close to one of the
partitions and far from the other one is sought. In a biological context, median and
closest SSPs relate to the identification of consensus
sequences or regions of similarity, a task of interest for creating
diagnostic probes for bacterial infections or discovering potential drug
targets \citep{Festa2007219} to give some examples. In this context, farthest
SSPs relate to, for example, the existence of hosts in which the
string identified is not preserved and hence should not be targeted.

Here we are specifically concerned with the \Problem{Far From Most
String Problem} (FFMSP), a problem which can be roughly described as
identifying a string whose distance is above a certain predefined threshold
for as many as possible strings in the input set. A formal definition of the
problem is provided in Section \ref{sec:problem:statement}. As pointed out in
\cite{Boucher2013107}, biological sequence data is subject to frequent random
mutations and errors and it can thus be problematic to force the
solution string to fit the entire input set. In this sense, the FFMSP provides
an interesting tradeoff since it tries to maximize the number of strings for
which the distance constraint is fulfilled, without thus requiring that this
is done for all strings in the input set. Not surprisingly, this turns out
to be a task of formidable difficulty (a summary of complexity results is
provided in Section  \ref{sec:computational:hardness}) so the use of complete
methods (exact or approximate) is often out of the question. Heuristic
methods are therefore required to solve this problem. See Section \ref{sec:related:work} for an
overview of related work in this direction.

We propose here a memetic approach to the FFMSP. Memetic algorithms (MAs)
\citep{bib:Moscato1989} are a broad class of metaheuristics that try to blend
together ideas from different optimization techniques, orchestrating the
interplay between global population-based search and local search components
-- see \cite{bib:HandbookMA2011,Neri20121} for recent treatments of MAs; we
also refer to \cite{bib:Ong2010Memetic} for an overview of the broad area of
memetic computing. Our MA is described in detail in Section \ref{sec:MA}. We
have conducted an extensive empirical evaluation of different variants of
this technique and --as shown in Section \ref{sec:results}-- it compares
favorably to previous approaches in the literature.

\section{The \Problem{Far From Most String Problem}}
In this section we formally state the FFMSP and analyze its complexity. We then go on to summarize relevant related work.

\subsection{Problem Statement}
\label{sec:problem:statement}
 Let $\Sigma$ denote a finite set of symbols (the alphabet), and let a string $s$ taken from such an alphabet be a finite sequence of zero or more of those symbols ($s \in \Sigma^*$).
We write $|s|$ for the length of  string $s$,
$s=s_1 s_2 \dots s_m$ ($s_i \in \Sigma, 1 \leqslant i \leqslant m$) to denote
string $s$ as a sequence of $m$ symbols ($s_i$ is $i$-th symbol of s),
 and $\epsilon$ for the empty string.

Let $s$ and $r$ be two finite strings of same length taken from same alphabet. The {\em Hamming Distance} (\HD) between those strings is the number of positions
at which their symbols are different:
\begin{equation}
\HD(s,r) = \sum_{1 \leqslant i \leqslant |s|}{ [s_i \neq r_i] }
\end{equation}
where $[]:\mathbb{B}\rightarrow\mathbb{N}$ is an indicator function ($[{\rm true}]=1$ and $[{\rm false}]=0$).

An instance of the FFMSP is given by a triple $(\Sigma, S, d)$, where
$\Sigma$ is a finite alphabet of symbols, $S = \{S^1, S^2, \dots, S^n\}$ is a finite set
of $n > 1$ strings taken from $\Sigma$ ($S^i \in \Sigma^m, 1 \leqslant i \leqslant n$), all of them with same length ($m$),
and $1 \leqslant d \leqslant m$
is an integer standing for the {\em distance threshold} of the problem.

A candidate solution $x$ for the problem  is a string of $m$ symbols
taken from $\Sigma$ ($x \in \Sigma^m$), and we say that $x$ is far from
string $S^i \in S$ if $\HD(x,S^i) \geqslant d$ (if $\HD(x,S^i) < d$ we say
that $x$ is near $S^i$). The objective function $f$ for a candidate
solution to the problem $x$
is the number of strings in $S$ that are far from $x$:
\begin{equation}
f(x) = \sum_{S^i \in S}{[\HD(x,S^i) \geqslant d]} \label{objective:function}
\end{equation}
and the FFMSP consists of maximizing such an objective function.

Note that the problem is trivial if $n < |\Sigma|$ as, in this case,
a string $x$ that is far from all strings in $S$ can be easily constructed
by taking for $x_i$ $(1 \leqslant i \leqslant |x|)$ a symbol that is not in position $i$ for any string in $S$.

\subsection{Computational Complexity of the FFMSP}
\label{sec:computational:hardness}

The FFMSP is NP-hard in general by reduction from \Problem{Independent Set}
(cf. \cite{KevinLanctot200341}). The proof is provided for $|\Sigma|=3$ and
can be generalized to $|\Sigma|\geqslant4$ with some changes. Of course, NP-hardness is a worst-case result arising from a unidimensional analysis. A
more careful multidimensional analysis may indicate the existence of a
structural \emph{parameter} whose value determines whether the problem's
resolution is tractable or not. Such an analysis is the subject of
parameterized complexity (see \cite{fellows02parameterized} for a general
overview). Therein, the notion of tractability is captured by class FPT (\emph
{fixed-parameter tractable}) \citep{Downey:1995:FTC:210940.210967}, comprising 
parameterized problems \PProblem{A}{k} for which a solution is computed in
time $O(f(k)n^c)$ where $n=|A|$, $k$ is the parameter, $f$ is an arbitrary
function of $k$ only, and $c$ is a constant independent of $k$. While the
FFMSP has not been directly addressed from the point of view of parameterized
complexity, the result in \cite{KevinLanctot200341} can be generalized to
this domain since the reduction used is a parameterized reduction (for most
purposes, a reduction between parameterized problems \PProblem{A}{k} and
\PProblem{B}{k'} is termed a \emph{parameterized reduction} if $B=\Phi(A, k)$, with $\Phi$ being computable in time $O(g(k)n^c)$, and $k'=h(k)$, where $g,h
$ are arbitrary functions of $k$ and $c$ is a constant independent of $k$
\citep{Downey:1995:FTC:210940.210967}). More precisely, the reduction in
\cite{KevinLanctot200341} is parameterized when the number of strings $k$ that must
be far from the constructed string is taken as a parameter of \Problem{FFMSP}
and the number $k'$ of vertices is the parameter in the \Problem{Independent
Set} problem. Unfortunately, \Problem{Independent Set} is W[1]-complete for
this parameterization \citep{Downey1995109} so the FFMSP turns out to be W[1]-
hard. This means that --unlike FPT problems-- for a fixed $k$ there only exist
algorithms running in time $O(f(k)n^{g(k)})$ with
$g(k) \rightarrow \infty$ \citep{Abrahamson1995235}.

Having established the general hardness of the FFMSP, attention can now be
turned to its approximability. In this sense, it is known that W[1]-hard
problems have no fully polynomial-time approximation scheme unless W[1]=FPT
\citep{Cai1997465}. The bad news does not end there since it has been also shown that 
\Problem{Independent Set} does not admit a PTAS \citep{Arora:1998:PVH:278298.278306},
and hence neither does the FFMSP. As a matter of fact, \Problem{Independent Set}
is known to be APX-hard --i.e. for some $\epsilon>0$ finding
a $(1+\epsilon)$-approximation is NP-hard-- in general and even for some
restricted cases \citep{Papadimitriou1991425}. Currently, it is known to be
inapproximable within a factor of $n/2^{(\log n)^{3/4+\epsilon}}$ unless NP
$\subseteq$ BPTIME($2^{(\log n)^{O(1)}}$) \citep{Khot06better}. Note, finally, 
that APX-hardness of \Problem{FFMSP} has been also established in the case
$|\Sigma|=2$ following a result of \cite{Boucher2013107}.

\subsection{Related Work}
\label{sec:related:work}
The FFMSP has attracted a lot of interest from the perspective of heuristics
due to the aforementioned hardness results. One of the first
heuristic attacks on the FFMSP was done by \cite{Meneses05optimization}.
They proposed a base heuristic which greedily picked symbols
to construct a string so as to maximize the number of strings whose distance
was above the threshold, and augmented it with a local search procedure based
on two-exchanges (which provided an additional 1-5\% improvement of the
results). Later on, \cite{Festa2007219} proposed a \emph{greedy
randomized adaptive search procedure} (GRASP) \citep{GRASP} approach. GRASP tries to
modulate the myopic behavior of greedy techniques by making randomized
choices from among the elements of a restricted candidate list (comprising the
best elements according to their greedy value). A local search component
analogous to that in \cite{Meneses05optimization} was used. Overall, this
approach provided improvements of 10-16\% using a similar computational time.

The GRASP in \cite{Festa2007219} has been benchmarked in several subsequent
publications. \cite{mousavi10hybridization} proposed a \emph{beam search} (BS) approach based on a heuristic objective function measuring the likelihood
of candidate solutions to lead to better solutions with as few changes as
possible. This BS approach was augmented with local search as well, and was
shown to outperform GRASP in over 80\% of test instances. The same objective
function was later used within the GRASP approach itself \citep{Mousavi12},
leading to huge improvements in solution quality. A genetic algorithm (GA)
was also proposed by \cite{festa12efficient}.
The main feature of this GA was the use of a diversity-management
policy by which the population was divided into three tiers: the first 
(tier A) comprising the best solutions in the previous generation,
the second (tier B) comprising the results of applying uniform crossover
to a parent from tier A and another parent of tiers B/C, and the third
(tier C) comprising randomly generated solutions. This GA provided average
improvements of 54-257\% over the previous GRASP, albeit at a higher
computational cost. Finally, \cite{FeroneFR13} proposed a
\emph{variable neighborhood search} (VNS) metaheuristic and its use both as a
stand-alone algorithm or within GRASP (as a local-search component). Also,
\emph{path relinking} is used as an intensification procedure, either in
GRASP, VNS or hybrids thereof. The GRASP+VNS+PR hybrid was shown to provide
the best results (sometimes with an up to a 5-fold increase with respect to
pure GRASP), although GRASP+PR offered better performance in a scenario of
limited computational budget.

\section{A Memetic Algorithm for the FFMSP}
\label{sec:MA}
As anticipated in Section \ref{sec:intro}, an MA is a metaheuristic approach based on the synergistic combination of local-search and population-based components. This is commonly achieved via the integration of an evolutionary algorithm (EA) and some local-search component(s), or in general via the intelligent interplay of different problem-specific
algorithms \citep{Moscato_NewMethods_99,moscato10modern}. In this sense, we have adopted an integrative perspective (cf. \citet{puchinger+:combining-iwinac05}), using an EA skeleton to which several complex algorithmic add-ons are plugged. \refFig{fig:MA} provides a pseudo-code for the proposed memetic algorithm. This MA
maintains a population of solutions to the problem (parameter $\mathit{popSize}$ corresponds to
the size of this population) that are firstly initialized and then recombined (with crossover
probability $p_{X}$), mutated (with mutation probability $p_m$)
and improved (by means of a local search procedure) until the allowed execution time is reached.
At the end, the best solution found is returned.

\begin{figure}[ht!]
     \begin{algorithm}{Memetic Algorithm}{}{}
     \LnFunction\kw{function} \fnctn{MA}$(I=(\Sigma, S = \{S^1, S^2, \dots, S^n\}, d),\alpha,\mathit{popSize},p_{X},p_m)$	
	   \LnNumber \kw{for} $i :=1$ \kw{to} $\mathit{popSize}$ \kw{do}
     \LnNumber \> $\mathit{pop}_i :=$ \fnctn{GRASP}$(\alpha,I)$
     \LnNumber \> \fnctn{Evaluate}($\mathit{pop}_i$)
     \LnNumber  \kw{end for}
     \LnNumber \kw{while}  allowed runtime \kw{not} exceeded  \kw{do}
		 \LnNumber \> \kw{if} recombination is performed($p_{X}$) \kw{then}
     \LnNumber \> \> $parent_1 :=$ \fnctn{Select}($\mathit{pop}$)
     \LnNumber \> \> $parent_2 :=$ \fnctn{Select}($\mathit{pop}$)
     \LnNumber \> \> $\mathit{offspring} :=$ \fnctn{Recombine}($parent_1$, $parent_2$)
     \LnNumber \> \kw{else}
     \LnNumber \> \> $\mathit{offspring}:=$ \fnctn{Select}($\mathit{pop}$)
     \LnNumber \> \kw{end if}
     \LnNumber \> $\mathit{offspring} :=$ \fnctn{Mutate}($p_m,\mathit{offspring}$)
     \LnNumber \> $\mathit{offspring}:=$ \fnctn{Local Search}($\mathit{offspring},I$)
     \LnNumber \> \fnctn{Evaluate}($\mathit{offspring}$)
     \LnNumber \> $\mathit{pop} :=$ \fnctn{Replace}($\mathit{pop}$, $\mathit{offspring}$)
     \LnNumber \kw{end while}	
     \LnNumber \kw{return} best solution found
     \LnFunction\kw{end function}
     \end{algorithm}
     \caption{Memetic Algorithm for FFMSP.}
     \label{fig:MA}
\end{figure}

\begin{figure}[ht!]
     \begin{algorithm}{Heuristic Function}{}{}
     \LnFunction\kw{function} $h(s,I=(\Sigma, S = \{S^1, S^2, \dots, S^n\}, d))$	
		 \LnNoNumber $m$ {\bf is} $|S^i|,\ \forall i\,.\,1 \leqslant i \leqslant n$
		 \LnNumber $\mathit{near} := 0$
	   \LnNumber \kw{for} $i := 1$ \kw{to} $\mathit{n}$ \kw{do}
     \LnNumber \> $d_i := \HD(s,S^i)$
     \LnNumber \> $c_i := m - d_i$
		 \LnNumber \> \kw{if} $d_i < d$ \kw{then}
		 \LnNumber \> \> $\mathit{near} :=  \mathit{near} + 1$
     \LnNumber \> \kw{end if}
     \LnNumber \kw{end for}
		 \LnNumber $\mathit{f} :=  n - \mathit{near}$
		 \LnNumber \kw{if} $\mathit{near} = 0$ \kw{then}
		 \LnNumber \> $\mathit{GpC} :=  0$
		 \LnNumber \kw{else}
		 \LnNumber \> $\mathit{sumGpC} :=  0$
	   \LnNumber \> \kw{for} $i := 1$ \kw{to} $\mathit{n},\ d_i < d$ \kw{do}
		 \LnNumber \> \> $g_i := 1$
	   \LnNumber \> \> \kw{for} $j := 1$ \kw{to} $\mathit{n},\ i \neq j$ \kw{do}
		 \LnNumber \> \> \> $\mathit{sumP} := 0$
	   \LnNumber \> \> \> \kw{for} $c := c_j$ \kw{to} $c_i$ \kw{do}
	   \LnNumber \> \> \> \> $\mathit{sumP} := \mathit{sumP} + T[c_i,c] / |\Sigma|^{c_i}$		
     \LnNumber \> \> \> \kw{end for}
		 \LnNumber \> \> \> $g_i := g_i + \mathit{sumP}$
     \LnNumber \> \> \kw{end for}
		 \LnNumber \> \> $\mathit{sumGpC} :=  \mathit{sumGpC} + g_i / c_i$
     \LnNumber \> \kw{end for}
		 \LnNumber \> $\mathit{GpC} :=  \mathit{sumGpC} / \mathit{near}$
     \LnNumber \kw{end if}
  	 \LnNumber \kw{return} $(n+1) \times f + \mathit{GpC}$
     \LnFunction\kw{end function}
     \end{algorithm}
     \caption{Heuristic function for FFMSP.}
     \label{fig:heuristic}
\end{figure}

Since the problem does not pose constraints
on the construction of solutions, these can be naturally represented as
strings of length $m$ over alphabet $\Sigma$. Although there exists a well-defined
objective function $f$ to be maximized, corresponding to the number of strings
in the instance that are far from the current solution (see \refEq{objective:function}),
we have instead considered the heuristic function proposed in \cite{Mousavi12} as it
reduces the number of local optima of the resulting search landscape. This heuristic
(which we henceforth refer to as $h$) evaluates a solution taking
into account its objective value but also
the {\em likelihood} of the solution to lead to better solutions with as few
local moves as possible. One important property of $h$ is that
$f(s) > f(s')$ implies $h(s) > h(s')$.
Details on the rationale of its design can be found
in \cite{Mousavi12}, and we reproduce its pseudo-code in \refFig{fig:heuristic}
for completeness of description of our proposal. In this pseudo-code,
$T$ is a bi-variable function from $\mathbb{N}_0 \times \mathbb{Z}$
to $\mathbb{N}_0$ that can be precomputed as follows:
$$
\begin{array}{lcl}
T(0,0) &=& 1\\
T(0,k) &=& 0, \mathrm{\ if\ } k \neq 0\\
T(L,k) &=& T(L-1,k-1) + (|\Sigma|-2)T(L-1,k) +  T(L-1,k+1), \mathrm{\ if\ } L > 0
\end{array}
$$
\noindent and, as a result, $h$ can be calculated in $O(nm+n^2)$, having paid a
one-off time cost of $O(m^2)$ to initialize a bi-dimensional array corresponding to $T$.

Initialization of individuals in the population was done by means of a
Greedy Randomized Adaptive Search Procedure (GRASP) similar
to the one proposed in \cite{FeroneFR13}. Pseudo-code for this procedure
is shown in \refFig{fig:GRASP}, where
$V_j(c)$  stands for the total number of occurrences of symbol
$c$ in position $j$ in any of the strings in the given instance, and
$V_j^{min}$ ($V_j^{max}$) is the number of occurrences of the least (most) frequent symbol at position $j$.
As can be seen, the procedure constructs a solution by adding a symbol to the
string on each iteration. This symbol is chosen uniformly at random
from a restricted candidate list (RCL) that includes candidate symbols. A parameter
$\alpha$ ($0 \leqslant \alpha \leqslant 1$) is introduced as a generalization of the original GRASP 
component of the algorithm by \citet{FeroneFR13}, in order to
control the degree of greediness used to generate 
the solution: smaller values for this parameter imply the generation of greedier
solutions and larger values introduce more randomness. For instance, if $\alpha$ is 0,
so is $\beta$, and then only the least frequent symbol is included in the RCL (this would
correspond to a greedy selection strategy where the least frequent symbol for each
position would be selected). For greater values of $\alpha$, more symbols can be included
in the RCL (all of them when $\beta = 1$), and then selection is more random.

\begin{figure}[ht!]
     \begin{algorithm}{GRASP Algorithm}{}{}
     \LnFunction\kw{function} $\fnctn{GRASP}(\alpha,I=(\Sigma, S = \{S^1, S^2, \dots, S^n\}, d))$
		 \LnNoNumber $m$ {\bf is} $|S^i|,\ \forall i\,.\, 1 \leqslant i \leqslant n$
		 \LnNumber $V_j(c) := \sum_{S^i \in S,\ S^i_j=c} 1,\ \forall c \in \Sigma,\  \forall j\,.\, 1 \leqslant j \leqslant m$
		 \LnNumber $V_j^{min} := \min \{ V_j(c)\,|\,c \in \Sigma \},\  \forall j\,.\, 1 \leqslant j \leqslant m$
		 \LnNumber $V_j^{max} := \max \{ V_j(c)\,|\,c \in \Sigma \},\  \forall j\,.\, 1 \leqslant j \leqslant m$
		 \LnNumber $s := \epsilon$
		 \LnNumber $\beta$ := \fnctn{Random}[0,$\alpha$]
		 \LnNumber \kw{for} $j := 1$ \kw{to} $\mathit{m}$ \kw{do}
     \LnNumber \> $\mathit{RCL}_j := \emptyset$
		 \LnNumber \> $\mu := V_j^{min} + \beta (V_j^{max} - V_j^{min})$
	   \LnNumber \> \kw{for} $c \in \Sigma,\ V_j(c) \leqslant \mu$ \kw{do}
     \LnNumber \> \> $\mathit{RCL}_j := {RCL}_j \cup \{c\}$
		 \LnNumber \> \kw{end for}
		 \LnNumber \> $s := s\ \fnctn{Random}(\mathit{RCL}_j)$
  	 \LnNumber \kw{end for}
  	 \LnNumber \kw{return}$(\mathit{s})$
     \LnFunction\kw{end function}
     \end{algorithm}
     \caption{GRASP Algorithm for FFMSP.}
     \label{fig:GRASP}
\end{figure}

Mutation of individuals has been implemented by substituting (with probability $p_m$) symbols in
current string with a random symbol from $\Sigma$, and recombination has been performed by means
of an intelligent operator that uses path relinking (PR) \citep{Glover00fundamentalsof} in order to attain a sensible
recombination of information from parents. PR finds new solutions by exploring paths that connect high quality
solutions in neighborhood space. This is achieved by starting a search in one of those solutions (the initiating one) and performing
local moves that lead to another solution (the guiding one). \refFig{fig:PR} shows the pseudo-code of this procedure
for combining two parent solutions ($p^1$ and $p^2$) for the FFMSP, where the worse parent (according to heuristic
function $h$) acts as the initiating solution ($s$) and the better
parent as the guiding one ($s^*$). As it can be seen,  differences between
both solutions (as their positions and symbols in $s^*$) are firstly computed (set $\Delta$),
thus defining the set of local moves to be done.
Then, a path from the initiating solution towards the guiding one is generated by incorporating
into the former, on each move, the component from the set $\Delta$ that leads to a better
intermediate solution (the incorporation of the selected component $(i^*,s^*)$
is done by replacing the symbol at position $i^*$ with $s^*$ in the current solution).
This procedure is repeated until all moves have been performed (i.e., the
guiding solution has been reached) and the best solution found along the path ($s^*$) is returned as a child.

\begin{figure}[ht!]
     \begin{algorithm}{Path Relinking Crossover Algorithm}{}{}
     \LnFunction\kw{function} $\fnctn{Path Relinking}(p^1, p^2, I=(\Sigma, S = \{S^1, S^2, \dots, S^n\}, d))$
		 \LnNoNumber $m$ {\bf is} $|S^i|,\ \forall i\,.\, 1 \leqslant i \leqslant n$
		 \LnNumber $s := \fnctn{arg\ min}\{ h(s,I) \ | \  s \in \{p^1,p^2\}\}$
		 \LnNumber $s^* := \fnctn{arg\ max}\{ h(s,I) \ | \  s \in \{p^1,p^2\}\}$
		 \LnNumber $h^* := h(s^*,I)$
		 \LnNumber $\Delta  := \{ (i,s^*_i) \ |\ i \in \{1 \dots m\}, s_i \neq s^*_i \}$
     \LnNumber \kw{while} $\Delta \neq \emptyset$ \kw{do}
     \LnNumber \> $(i^*,c^*) := \fnctn{arg\ max}\{ h(s_1 s_2 \dots s_{i-1}\, c \ s_{i+1} \dots s_m,I) \ |\ (i,c) \in \Delta \}$
		 \LnNumber \> $s := s_1 s_2 \dots s_{i^*-1}\, c^* \ s_{i^*+1} \dots s_m$
		 \LnNumber \> $\Delta := \Delta - \{ (i^*,c^*) \}$
		 \LnNumber \> \kw{if} $h(s,I) > h^*$ \kw{then}
		 \LnNumber \> \> $s^* := s$
		 \LnNumber \> \> $h^* := h(s,I)$
  	 \LnNumber \> \kw{end if}
  	 \LnNumber \kw{end while}
  	 \LnNumber \kw{return}$(s^*)$
     \LnFunction\kw{end function}
     \end{algorithm}
     \caption{Path Relinking Crossover Algorithm for FFMSP.}
     \label{fig:PR}
\end{figure}

\begin{figure}[ht!]
     \begin{algorithm}{Local Search Algorithm}{}{}
     \LnFunction\kw{function} $\fnctn{Local Search}(s, I=(\Sigma, S = \{S^1, S^2, \dots, S^n\}, d))$
		 \LnNoNumber $m$ {\bf is} $|S^i|,\ \forall i\,.\, 1 \leqslant i \leqslant n$
		 \LnNumber $\mathit{improvement} := \textrm{true}$
     \LnNumber \kw{while} $\mathit{improvement}$ \kw{do}
		 \LnNumber \> $\mathit{improvement} := \textrm{false}$
     \LnNumber \> \kw{for} $j := 1$ \kw{to} $\mathit{m}$ \kw{do}
  	 \LnNumber \> \> \kw{for} $c \in \Sigma,\ c \neq s_j$ \kw{do}
		 \LnNumber \> \> \> $s' := s_1 s_2 \dots s_{j-1}\, c\ s_{j+1} \dots s_m$
		 \LnNumber \> \> \> \kw{if} $h(s',I) > h(s,I)$ \kw{then}
		 \LnNumber \> \> \> \> $s := s'$
		 \LnNumber \> \> \> \> $\mathit{improvement} := \textrm{true}$
		 \LnNumber \> \> \> \kw{end if}  	
     \LnNumber \> \> \kw{end for}
  	 \LnNumber \> \kw{end for}
  	 \LnNumber \kw{end while}
  	 \LnNumber \kw{return}$(\mathit{s})$
     \LnFunction\kw{end function}
     \end{algorithm}
     \caption{Local Search Algorithm for FFMSP.}
     \label{fig:HC}
\end{figure}

Finally, improvement of solutions is done through a sequential hill climbing
local search procedure on the search space induced by $h$ heuristic
function (see \refFig{fig:HC}), where a local move corresponds to modifying a single symbol in the solution
with another one from the alphabet.

\section{Results and Discussion}
\label{sec:results}

In order to evaluate the different heuristics, we have considered two benchmark sets.
Instances in the first set -- henceforth referred to as \textsc{RandomSet} -- are composed of different numbers ($n$) of strings
of the same length ($m$) randomly selected according to a uniform distribution
from the alphabet $\Sigma = \{\mathrm{A},\mathrm{T},\mathrm{C},\mathrm{G}\}$. Eighteen subsets of instances
have been defined using
different numbers of strings ($n \in \{100,200\}$),
string lengths ($m \in \{300, 600, 800\}$)
and distance thresholds ($d \in \{0.75 \cdot m, 0.80 \cdot m, 0.85 \cdot m\}$). For each of the eighteen subsets five different
instances have been generated yielding thus a total of 90 instances.

The second set  -- referred to as \textsc{RealSet} -- is also composed of 90 instances with the same combination of parameters as those
in \textsc{RandomSet}, but instead of being randomly generated, different strings in each instance have been selected from a
random segment of a real genome, namely Phytophthora Ramorum's genome \citep{PhytophthoraRamorum} (available at
\url{http://genome.jgi.doe.gov/Phyra1_1/Phyra1_1.download.html}).

Regarding the programs used in the experiments, all algorithms were coded in C and compiled using gcc under Linux
(the hybrid GRASP+VNS+PR by \cite{FeroneFR13} was available from their authors and we coded
remaining algorithms). With the purpose of making a fair comparison, all executions for different
algorithms were performed on the same machine (HP Proliant SL170s computer, Intel Xeon X5660 2.8 GHz processor, 8GB RAM,
CentOS 5.5 operating system) with a time limit of 600 seconds per execution. In all experiments, crossover probability was set to $p_{X}=0.9$,
mutation probability was $p_m=1/m$ ($m$ stands for the length of strings in the instance),
population size was $\mathit{popSize} = 100$ and binary tournament was used to perform selection.

In order to report results in tables and figures uniformly, we calculate the relative percentage
distance (RPD) of the solution obtained by each algorithm from the best result for the corresponding instance
obtained by any of the compared algorithms (defined as $(\subindex{sol}{best} - sol) \times 100 / \subindex{sol}{best}$),
and provide statistical values for those distances.

The statistical analysis of the results of the experiments was done using software available on the
web page of Research Group Soft Computing and Intelligent
Information Systems at the University of Granada (\url{http://sci2s.ugr.es/sicidm}).
With the aim of analyzing the statistical significance of the results, we used the following
methodology: firstly, we performed an Aligned Friedman Rank Test \citep{hodgesjr1962rmc},
a multiple-comparison non-parametric test that aims to detect significant
differences between the behavior of two or more algorithms and then ranks them from the 
best to worst. If, as a result of this test, the null hypothesis stating equality of
rankings between the populations is rejected, we proceed to post-hoc procedures in order
to compare the control algorithm (the best performing one) with the remaining algorithms.
The post-hoc procedures that we have considered are Bonferroni--Dunn's \citep{Dunn1961},
Holm's \citep{Holm1979}, Hochberg's\citep{Hochberg1988},
Hommel's \citep{Hommel1988}, Holland's \citep{HollandCopenhaver1987}, Rom's \citep{Rom1990},
Finner's \citep{Finner1993} and Li's \citep{Li2008} procedures. Finally, we report
adjusted $p$-values for different post-hoc procedures
(the smaller overall significance level at which the particular
null hypothesis
stating equality between the distributions obtained by the control algorithm
and the other compared algorithm would be rejected).

\subsection{Parameterization and Sensitivity Analysis}

So as to analyze the influence of the different algorithmic components and parameters
in the performance of the proposed MA, we firstly compared different variants of the algorithm. For these
experiments, we used instances in \textsc{RandomSet}, and performed 10 independent executions
for each algorithm and instance (thus, 50 executions per subset and algorithm).

The different components and parameters of the MA that we took into account were:
\begin{itemize}

\item Population initialization: we considered different values for parameter $\alpha$
in the GRASP algorithm (\refFig{fig:GRASP}) used during the initialization of the MA population
($\alpha \in \{0.1, 0.25, 0.5\}$). This parameter controls the greediness of GRASP, so
that greater values imply less greediness. We also considered completely randomly initializing individuals.

\item Recombination operator: we considered two possibilities; either using uniform crossover (UX)
or using a path relinking (PR) algorithm (\refFig{fig:PR}).

\item Local search operator: we considered the possibility of using the hill climbing operator (\refFig{fig:HC})
after mutation in the MA loop or not using local search at all.
\end{itemize}

In order to name different algorithms, term GRASP$_p$ indicates
using GRASP for initializing the population and
setting $\alpha=p$ in this
component of the algorithm. If the population was initialized randomly, the name of the
algorithm instead includes the term RND. If the name includes the term PR, then path relinking was used as the crossover operator.
Term UX denotes the use of uniform crossover and term HC indicates that the algorithm uses hill
climbing as the local search operator. According to this convention, \maPRHC{0.25}, for instance, is a
memetic algorithm that uses path relinking and hill climbing and sets $\alpha=0.25$ in the GRASP
algorithm for initializing the population.  \maUXRND\ corresponds to using uniform crossover, no local search and random
initialization of population (note that this particular algorithm is not a memetic algorithm but
a genetic algorithm).

\begin{figure}[!ht]
\begin{center}
\makebox[\textwidth][c]{\includegraphics[angle=0,scale=0.3]{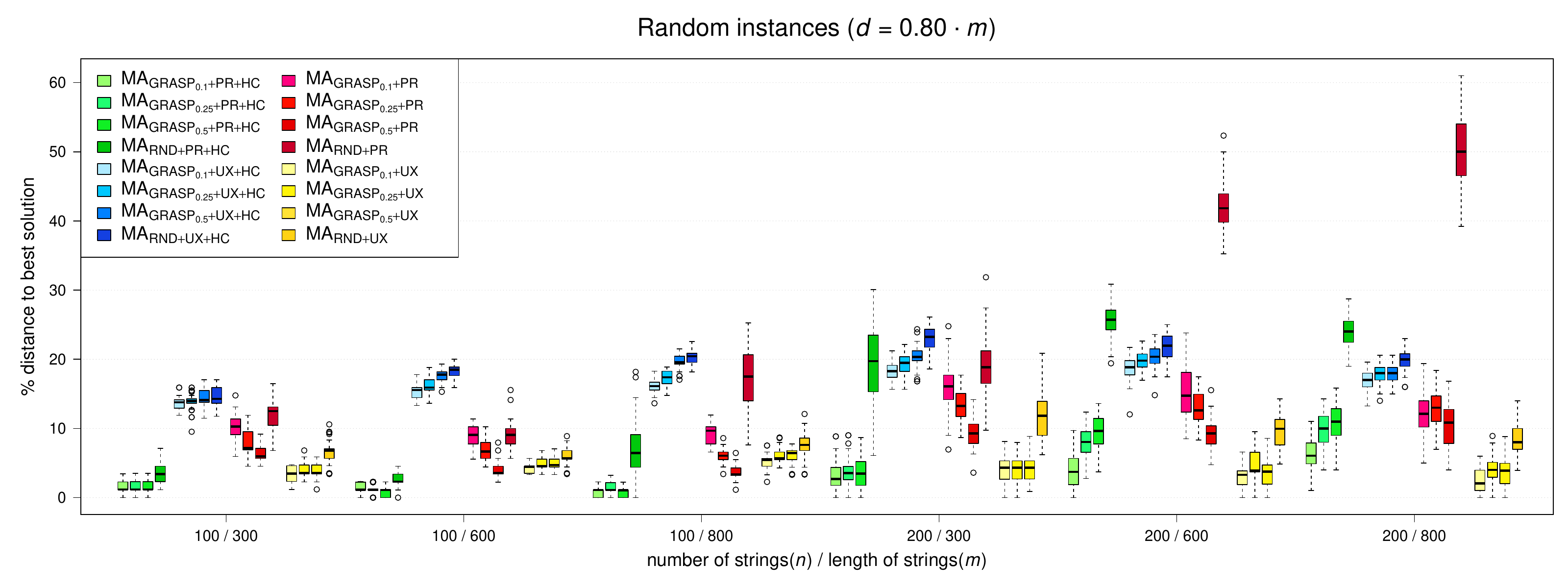}}
\end{center}
\caption{Box plots for relative percentage distances (RPD) from best solutions
of results obtained by each memetic algorithm for instances in \textsc{RandomSet}
and distance threshold $d=0.80 \cdot m$.
We have considered 6 datasets, each one comprising 5 different instances. Instances on each dataset are
characterized by a number of strings
($n$) of the same length ($m$) and a distance threshold ($d$). For each combination of $n/m$ parameters, box plots
in left-to-right order correspond to algorithms in legend in top-to-bottom, left-to-right order.}
\label{fig:mas:080}
\end{figure}

\begin{figure}[!ht]
\begin{center}
\makebox[\textwidth][c]{\includegraphics[angle=0,scale=0.3]{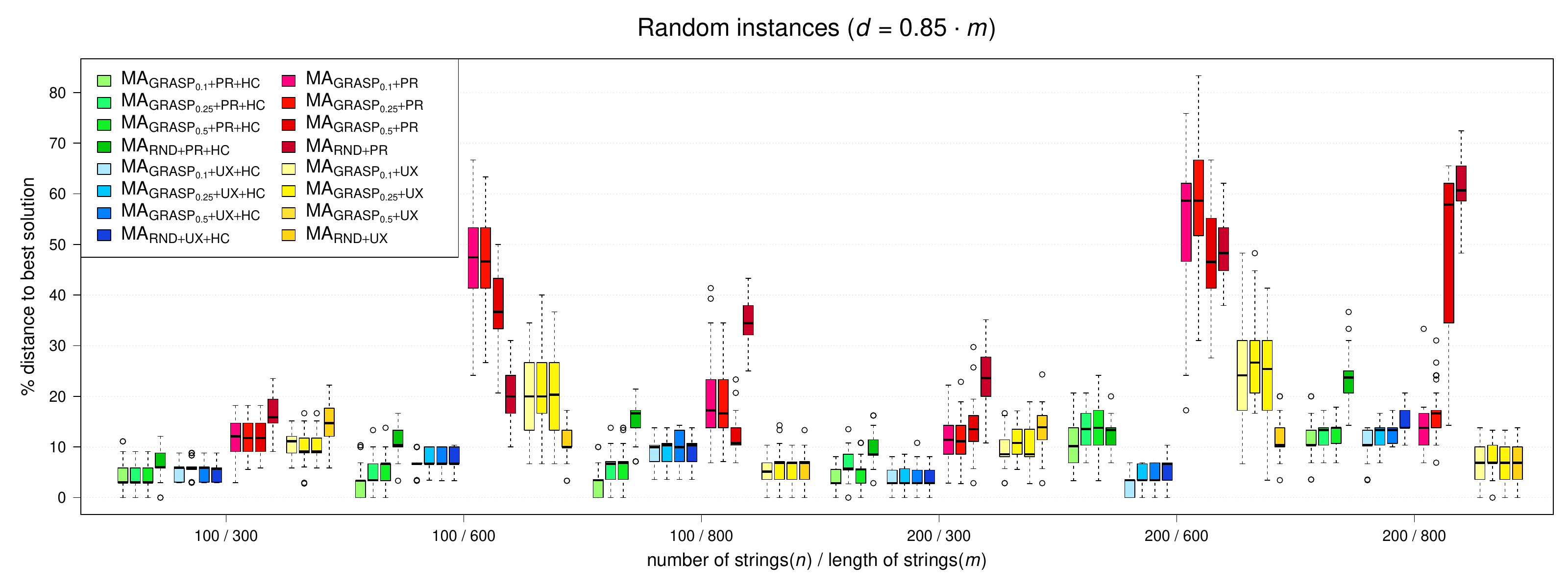}}
\end{center}
\caption{Box plots for relative percentage distances (RPD) from best solutions
of results obtained by each memetic algorithm for instances in \textsc{RandomSet}
and distance threshold $d=0.85 \cdot m$.
We have considered 6 datasets, each one comprising 5 different instances. Instances on each dataset are
characterized by a number of strings
($n$) of the same length ($m$) and a distance threshold ($d$). For each combination of $n/m$ parameters, box plots
in left-to-right order correspond to algorithms in legend in top-to-bottom, left-to-right order.}
\label{fig:mas:085}
\end{figure}

\refFig{fig:mas:080} shows the results of these experiments as boxplots for the RPD from best solutions
of results obtained by each algorithm on each subset for different numbers of strings and string lengths when
the distance threshold is $d=0.80 \cdot m$. It can be seen that, in most of the subsets, the memetic algorithms
using GRASP with path relinking and hill climbing obtain best results, and overall, a small value for $\alpha$
leads to better quality solutions. When $n=200$ and $m \geqslant 600$, algorithms using GRASP with uniform crossover and
no local search obtain best results. 
This may be due to the comparatively higher computational cost of PR and HC in these particular instances, which may consume the CPU budget faster (note at any rate that the difference is small with respect to \MA). This is further vindicated by the comparatively worse results of UX in combination with HC, indicating the interplay of the local improvement operator is better with PR (a more intensive recombination operator, capable of efficiently locating regions of interest) than with UX (a purely explorative, information-mixing operator).
For $d=0.85 \cdot m$, results are similar (\refFig{fig:mas:085}), but
best solutions are obtained by algorithms using GRASP with uniform crossover and
hill climbing when $n=200$ and $m=600$ (these algorithms also obtain results as good as those
obtained using path relinking when $n=200$ and $m=300$ or $m=800$, but perform worse in other cases).
No differences could be observed in the case of $d=0.75 \cdot m$, as all algorithms were able to obtain
optimal solutions in all runs.

To detect statistical significances in
the performance of different memetic algorithms,
we performed an Aligned Friedman Rank Test, with the results shown in
\refTable{friedman:random:MAs}  (distributed
according to $\chi^2$ with 15 degrees of freedom: 844.06). The $p$-value calculated by this test
is $p=2.44\cdot10^{-10}$, and this provides strong evidence for rejecting the null hypothesis that states equality of
rankings between the populations. It can also be observed that
\maPRHC{0.1} is ranked as the best performing algorithm followed by \maPRHC{0.25} and \maPRHC{0.5}.
As the Aligned Friedman Rank Test found statistical differences between the different
algorithms, we proceeded to perform post-hoc procedures, comparing
 the control algorithm (\maPRHC{0.1}) with those remaining.
\refTable{friedman:random:adjusted:p:MAs} shows adjusted $p$-values for
the different procedures. The null hypothesis
that states equality between the distributions obtained by the control algorithm
and the other compared algorithm was rejected by all procedures for all algorithms
(results are statistically significant) except for \maPRHC{0.25} and \maPRHC{0.5}.
In the case of  \maPRHC{0.5}, all procedures except for Bonferroni--Dunn's reject null
hypothesis (with $p$-values equal to 0.02 and 0.04, thus differences are statistically
significant). Finally, for \maPRHC{0.25}, none of the procedures found statistical
evidence to reject the null hypothesis, thus the performance of this algorithm cannot be considered
statistically different from the one of \maPRHC{0.1}.

\begin{table}[!ht]
\small \centering
\caption{Results for the Aligned Friedman Rank Test on \textsc{RandomSet} instances for different memetic algorithms.}
\begin{tabular}{rlr}
position & algorithm & ranking\\
\hline
 1\textsuperscript{st} & \maPRHC{0.1}  & 4125.76\\
 2\textsuperscript{nd} & \maPRHC{0.25} & 4482.05\\
 3\textsuperscript{rd} & \maPRHC{0.5}  & 4584.02\\
 4\textsuperscript{th} & \maUX{0.1}    & 5591.35\\
 5\textsuperscript{th} & \maUX{0.5}    & 5699.21\\
 6\textsuperscript{th} & \maUX{0.25}   & 5851.03\\
 7\textsuperscript{th} & \maUXRND    & 6206.48\\
 8\textsuperscript{th} & \maPR{0.5}    & 7615.16\\
 9\textsuperscript{th} & \maPRHCRND  & 7997.19\\
10\textsuperscript{th} & \maUXHC{0.1}  & 8206.66\\
11\textsuperscript{th} & \maUXHC{0.25}  & 8473.52\\
12\textsuperscript{th} & \maUXHC{0.5}  & 8504.02\\
13\textsuperscript{th} & \maPR{0.25}    & 8569.03\\
14\textsuperscript{th} & \maUXHCRND  & 8722.04\\
15\textsuperscript{th} & \maPR{0.1}  & 9446.67\\
16\textsuperscript{th} & \maPRRND  & 11133.78\\
\hline
\end{tabular}
\label{friedman:random:MAs}
\end{table}

\begin{table}[!ht]
\small \centering
\tabcolsep=0.06cm
\caption{Adjusted $p$-values for $N \times 1$ comparisons of control algorithm (\maPRHC{0.1}) with
different memetic algorithms for the Aligned Friedman Rank Test on \textsc{RandomSet} instances.}
\makebox[\textwidth][c]{
\resizebox{\textwidth}{!}{
\begin{tabular}{lrrrrrrrr}
algorithm&Bonf&Holm&Hoch&Homm&Holl&Rom&Finn&Li\\
\hline
\maPRHC{0.25}&1.00E0&7.00E-2&7.00E-2&7.00E-2&7.00E-2&7.00E-2&7.00E-2&7.00E-2\\
\maPRHC{0.5}&2.90E-1&4.00E-2&4.00E-2&4.00E-2&4.00E-2&4.00E-2&2.00E-2&2.00E-2\\
\maUX{0.1}&1.13E-12&2.25E-13&2.25E-13&2.25E-13&2.25E-13&2.25E-13&8.66E-14&8.06E-14\\
\maUX{0.5}&1.47E-14&3.93E-15&3.93E-15&3.93E-15&4.00E-15&3.74E-15&1.22E-15&1.05E-15\\
\maUX{0.25}&1.98E-17&6.61E-18&6.61E-18&6.61E-18&0.0&6.28E-18&0.0&1.42E-18\\
\maUXRND&3.70E-25&1.48E-25&1.48E-25&1.48E-25&0.0&1.41E-25&0.0&2.65E-26\\
\maPR{0.5}&9.50E-70&4.43E-70&4.43E-70&4.43E-70&0.0&4.21E-70&0.0&6.80E-71\\
\maPRHCRND&1.08E-85&5.74E-86&5.74E-86&5.74E-86&0.0&5.46E-86&0.0&7.71E-87\\
\maUXHC{0.1}&3.89E-95&2.34E-95&2.34E-95&2.34E-95&0.0&2.22E-95&0.0&2.79E-96\\
\maUXHC{0.25}&6.99E-108&4.66E-108&4.66E-108&4.66E-108&0.0&4.43E-108&0.0&5.01E-109\\
\maUXHC{0.5}&2.17E-109&1.59E-109&1.59E-109&1.59E-109&0.0&1.51E-109&0.0&1.55E-110\\
\maPR{0.25}&1.22E-112&9.78E-113&9.78E-113&9.78E-113&0.0&9.30E-113&0.0&8.76E-114\\
\maUXHCRND&1.78E-120&1.55E-120&1.55E-120&1.55E-120&0.0&1.47E-120&0.0&1.28E-121\\
\maPR{0.1}&3.58E-161&3.34E-161&3.34E-161&3.34E-161&0.0&3.17E-161&0.0&2.56E-162\\
\maPRRND&6.60E-279&6.60E-279&6.60E-279&6.60E-279&0.0&6.27E-279&0.0&4.73E-280\\
\hline
\end{tabular}
}
}
\label{friedman:random:adjusted:p:MAs}
\end{table}

As a result of this analysis, it can be concluded that incorporating
GRASP (with an small setting for parameter $\alpha$) in order to initialize
the population, using path relinking as a crossover operator and performing
local search by means of hill climbing yields overall best results and thus
using these components in the memetic algorithm is beneficial.

\subsection{Comparison to state-of-the-art algorithms}

In this section we compare one of the best performing memetic algorithm (\maPRHC{0.1} -- denoted by \MA\ henceforth for simplicity)
to state-of-the-art algorithms for the FFMSP (hybrid GRASP+VNS+PR in \cite{FeroneFR13} -- that we denote by \GRASPFFR --
and GRASP algorithm in \cite{Mousavi12} -- denoted by \GRASPMou). 
For \GRASPFFR\ we set corresponding parameters as indicated by the authors of that algorithm, and for \GRASPMou\ we set $\gamma=0.95$, after observing that using this setting provides better results than using the one suggested by the authors of this algorithm.
We firstly made a comparison on instances in \textsc{RandomSet},
for which 20 independent runs on each of the instances were executed by each
of the algorithms (thus, 100 executions per subset and algorithm). Results are shown numerically in
\refTable{tab:random} as the mean objective value obtained by each of the algorithms
on each subset, together with the mean and standard deviation for RPD for best solutions and mean improvement
percentage of results obtained by \MA\ with respect to those obtained by \GRASPFFR\ and \GRASPMou.

\begin{table}[!ht]
\small
\caption{Results obtained by \MA, \GRASPFFR\ and \GRASPMou\ for instances in \textsc{RandomSet}. We have considered
18 datasets, each one comprising 5 different instances. Instances on each dataset are characterized by a number of strings
($n$) of the same length ($m$) and a distance threshold ($d$). Table shows mean solution obtained by each algorithm (sol),
along with statistical values (mean ($\mu$) and standard deviation ($\sigma$)) for relative percentage distance (RPD)%
from best solutions of results obtained by each algorithm. For \GRASPFFR\ and \GRASPMou\ algorithms, mean improvement
percentage of solutions (imp.\%) obtained by \MA\ with respect to the ones obtained by these algorithms is also shown.}
\tabcolsep=0.06cm
\makebox[\textwidth][c]{
\resizebox{\textwidth}{!}{
\begin{tabular}{rrrrrrrrrrrrrr}
& & & & \multicolumn{2}{c}{\MA} && \multicolumn{3}{c}{\GRASPFFR} && \multicolumn{3}{c}{\GRASPMou}\\
            \cline{5-6} \cline{8-10} \cline{12-14}
$d$ & $n$ & $m$ && sol. & RPD $\mu \pm \sigma$ && sol. & RPD $\mu \pm \sigma$ & imp.\% && sol. & RPD $\mu \pm \sigma$ & imp.\% \\	
\cline{1-3} \cline{5-6} \cline{8-10} \cline{12-14}
$0.75 \cdot m$  & 100 & 300 && 100.00 & 0.00 $\pm$ 0.00 && 100.00 &  0.00 $\pm$ 0.00 &  0.00 && 100.00 &  0.00 $\pm$  0.00 &  0.00\\
                & 100 & 600 && 100.00 & 0.00 $\pm$ 0.00 && 100.00 &  0.00 $\pm$ 0.00 &  0.00 && 100.00 &  0.00 $\pm$  0.00 &  0.00\\
                & 100 & 800 && 100.00 & 0.00 $\pm$ 0.00 && 100.00 &  0.00 $\pm$ 0.00 &  0.00 && 100.00 &  0.00 $\pm$  0.00 &  0.00\\
                & 200 & 300 && 200.00 & 0.00 $\pm$ 0.00 && 200.00 &  0.00 $\pm$ 0.00 &  0.00 && 200.00 &  0.00 $\pm$  0.00 &  0.00\\
                & 200 & 600 && 200.00 & 0.00 $\pm$ 0.00 && 200.00 &  0.00 $\pm$ 0.00 &  0.00 && 200.00 &  0.00 $\pm$  0.00 &  0.00\\
                & 200 & 800 && 200.00 & 0.00 $\pm$ 0.00 && 200.00 &  0.00 $\pm$ 0.00 &  0.00 && 200.00 &  0.00 $\pm$  0.00 &  0.00\\
\cline{1-3} \cline{5-6} \cline{8-10} \cline{12-14}
$0.80 \cdot m$  & 100 & 300 &&  84.82 & 1.14 $\pm$ 0.75 &&  80.78 &  5.85 $\pm$ 1.51 &  4.71 &&  70.99 & 17.27 $\pm$  1.65 & 16.13\\
                & 100 & 600 &&  87.08 & 1.27 $\pm$ 0.72 &&  79.12 & 10.30 $\pm$ 1.47 &  9.03 &&  70.83 & 19.69 $\pm$  1.06 & 18.42\\
                & 100 & 800 &&  89.90 & 1.21 $\pm$ 0.81 &&  79.52 & 12.60 $\pm$ 1.75 & 11.40 &&  71.08 & 21.88 $\pm$  0.97 & 20.68\\
                & 200 & 300 && 109.58 & 2.34 $\pm$ 1.49 && 105.85 &  5.66 $\pm$ 2.26 &  3.32 &&  83.04 & 25.98 $\pm$  1.69 & 23.64\\
                & 200 & 600 && 101.23 & 2.86 $\pm$ 1.91 &&  88.95 & 14.61 $\pm$ 2.60 & 11.76 &&  80.90 & 22.29 $\pm$  2.65 & 19.44\\
                & 200 & 800 &&  93.82 & 4.26 $\pm$ 2.28 &&  80.09 & 18.27 $\pm$ 3.26 & 14.01 &&  79.77 & 18.59 $\pm$  1.97 & 14.33\\
\cline{1-3} \cline{5-6} \cline{8-10} \cline{12-14}
$0.85 \cdot m$  & 100 & 300 &&  32.58 & 3.61 $\pm$ 2.28 &&  18.41 & 45.60 $\pm$ 5.55 & 41.99 &&  30.10 & 10.95 $\pm$  1.61 &  7.34\\
                & 100 & 600 &&  28.76 & 4.76 $\pm$ 3.12 &&   4.89 & 83.82 $\pm$ 3.38 & 79.06 &&  25.36 & 16.01 $\pm$  2.12 & 11.25\\
                & 100 & 800 &&  27.96 & 2.90 $\pm$ 2.61 &&   2.58 & 91.03 $\pm$ 3.25 & 88.13 &&  24.33 & 15.51 $\pm$  1.70 & 12.61\\
                & 200 & 300 &&  34.49 & 3.64 $\pm$ 2.29 &&  14.85 & 58.57 $\pm$ 4.34 & 54.93 &&  32.69 &  8.62 $\pm$  2.80 &  4.98\\
                & 200 & 600 &&  26.17 & 6.54 $\pm$ 4.39 &&   2.26 & 91.93 $\pm$ 2.60 & 85.39 &&  25.54 &  8.76 $\pm$  2.84 &  2.22\\
                & 200 & 800 &&  25.61 & 4.42 $\pm$ 2.95 &&   0.60 & 97.76 $\pm$ 1.83 & 93.35 &&  23.71 & 11.51 $\pm$  2.97 &  7.10\\
\cline{1-3} \cline{5-6} \cline{8-10} \cline{12-14}
\end{tabular}
}
}
\label{tab:random}
\end{table}

\begin{figure}[!ht]
\begin{center}
\includegraphics[angle=0,scale=0.4]{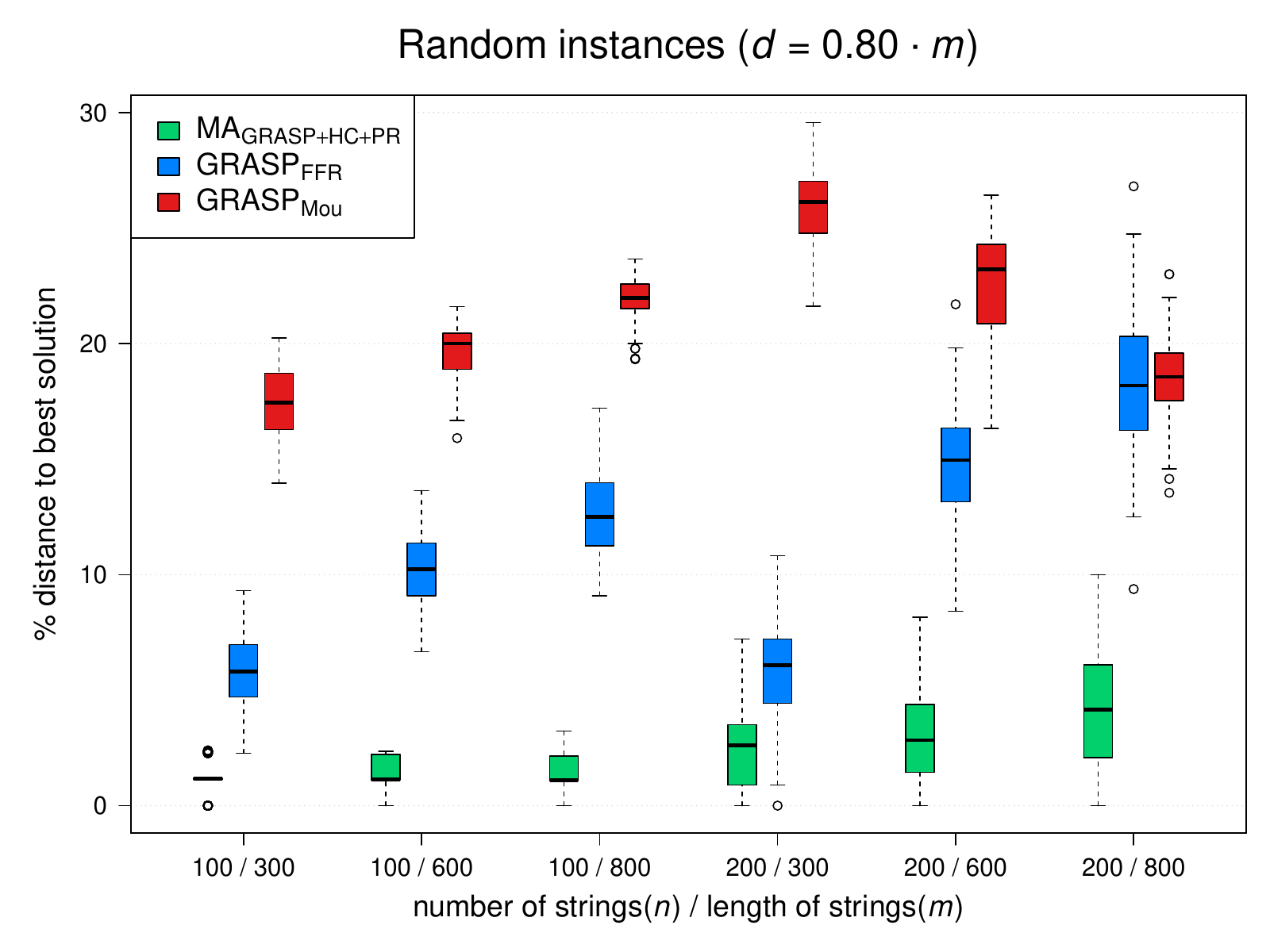}
\end{center}
\caption{Box plots for relative percentage distances (RPD) from best solutions
of results obtained by state-of-the-art algorithms for the FFMSP (\GRASPFFR\ and \GRASPMou) and \MA\
for instances in \textsc{RandomSet}
and distance threshold $d=0.80 \cdot m$.
Instances on each dataset are
characterized by a number of strings
($n$) of the same length ($m$) and a distance threshold ($d$). For each combination of $n/m$ parameters, box plots
in left-to-right order correspond to algorithms in legend in top-to-bottom order.}
\label{fig:random:080}
\end{figure}

For instances with $d = 0.75 \cdot m$, all the algorithms were always able to find optimal
solutions on all instances. In the case of $d = 0.80 \cdot m$, the best results were obtained
by \MA\ followed by \GRASPFFR. The average improvement percentage of \MA\ over \GRASPFFR\ ranges from
3.32\% for $n=200$ and $m=300$ to 14.01\% for $n=200$ and $m=800$. The mean average
improvement of \MA\ over \GRASPFFR\ is 9.04\% whereas 
it is 18.77\% for \MA\ over \GRASPMou. \refFig{fig:random:080} shows the results of these experiments as boxplots for the RPD from the best solutions
of results obtained by each algorithm on each subset. As can be seen, the average improvement percentage of
\MA\ over \GRASPFFR\ increases with $m$, both for $n=100$ and $n=200$. In the case of  \GRASPMou, this percentage is rather stable for $n=100$ and seems to decrease gently for increasing $m$ with $n=200$, but is always larger than for \GRASPFFR. For instances with $d = 0.85 \cdot m$,
the best results are again obtained by \MA\ followed by \GRASPMou. The average improvement percentage of
\MA\ over \GRASPMou\ ranges from 2.22\% for $n=200$ and $m=600$ to 12.61\% for $n=100$ and $m=800$.
The mean average
improvement of \MA\ over \GRASPMou\ is 7.58\% whereas for \MA\ over \GRASPFFR\
it is 73.81\%.

\begin{figure}[!ht]
\begin{center}
\includegraphics[angle=0,scale=0.4]{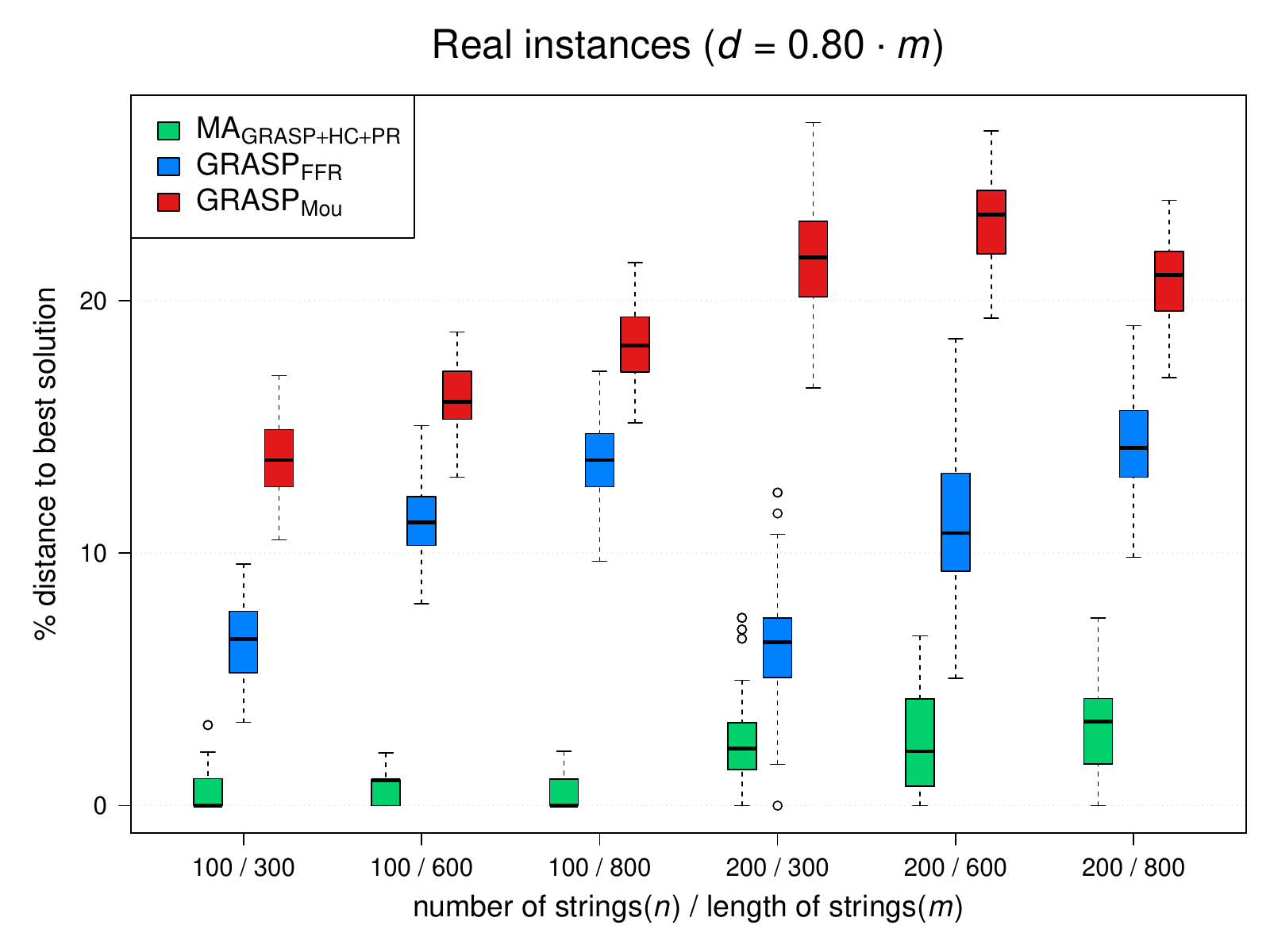}
\end{center}
\caption{Box plots for relative percentage distances (RPD) from best solutions
of results obtained by state-of-the-art algorithms for the FFMSP (\GRASPFFR\ and \GRASPMou) and \MA\
for instances in \textsc{RealSet}
and distance threshold $d=0.80 \cdot m$.
Instances on each dataset are
characterized by a number of strings
($n$) of the same length ($m$) and a distance threshold ($d$). For each combination of $n/m$ parameters, box plots
in left-to-right order correspond to algorithms in legend in top-to-bottom order.}
\label{fig:real:080}
\end{figure}

\refTable{friedman:random} shows statistical results for an Aligned Friedman Rank Test
(distributed
according to $\chi^2$ with 2 degrees of freedom: 1327.33). The $p$-value calculated by this test
is 0, and this indicates strong evidence for rejecting the null hypothesis that states equality of
rankings between the populations. We can also observe that \MA\ is ranked as the best performing
algorithm followed by \GRASPMou\ and \GRASPFFR.

\begin{table}[!ht]
\small \centering
\caption{Results for the Aligned Friedman Rank Test on \textsc{RandomSet} instances
for \MA\ and different state-of-the-art algorithms.}
\begin{tabular}{rlr}
position & algorithm & ranking\\
\hline
1\textsuperscript{st} & \MA       & 1471.59\\
2\textsuperscript{nd} & \GRASPMou & 2998.60\\
3\textsuperscript{rd} & \GRASPFFR & 3631.29\\
\hline
\end{tabular}
\label{friedman:random}
\end{table}

Since the Aligned Friedman Rank Test found statistical differences between the different
algorithms, we proceeded to perform post-hoc procedures in order
to compare the control algorithm (\MA) with other algorithms (\GRASPMou\ and \GRASPFFR).
In all cases and with all procedures, the null hypothesis
that states equality between the distributions obtained by the control algorithm
and the other compared algorithm was rejected, thus there are statistical
differences between the control algorithm and those remaining. The adjusted $p$-values
for different post-hoc procedures are shown in \refTable{friedman:random:adjusted:p}. These $p$-values are 0 or very close to 0 and hence the results are considered  statistically significant. 

\begin{table}[!ht]
\small \centering
\tabcolsep=0.06cm
\caption{Adjusted $p$-values for $N \times 1$ comparisons of control algorithm (\MA) with
different state-of-the-art algorithms for the Aligned Friedman Rank Test on \textsc{RandomSet} instances.}
\makebox[\textwidth][c]{
\resizebox{\textwidth}{!}{
\begin{tabular}{lrrrrrrrr}
algorithm&Bonf&Holm&Hoch&Homm&Holl&Rom&Finn&Li\\
\hline
\GRASPFFR&0.0      &0.0      &0.0      &0.0      &0.0      &0.0      &0.0&0.0\\
\GRASPMou&1.73E-189&8.66E-190&8.66E-190&8.66E-190&0.0      &8.66E-190&0.0&8.66E-190\\
\hline
\end{tabular}
}
}
\label{friedman:random:adjusted:p}
\end{table}

For instances in \textsc{RealSet} we also executed 20 independent runs by each
of the algorithms on each instance (thus, 100 executions per subset and algorithm). 
The results are also shown numerically in
\refTable{tab:real} as the mean objective value obtained by each of the algorithms
on each subset, together with the mean and standard deviation for RPD for best solutions and mean improvement
percentage of solutions obtained by \MA\ with respect to those obtained by \GRASPFFR\ and \GRASPMou.

The results are very similar to those on \textsc{RandomSet} instances.
For instances with $d = 0.75 \cdot m$, all algorithms were always able to find optimal
solutions on all instances. In the case of $d = 0.80 \cdot m$, the best results were obtained
by \MA\ followed by \GRASPFFR. The average improvement percentage of \MA\ over \GRASPFFR\ ranges from
4.06\% for $n=200$ and $m=300$ to 13.14\% for $n=100$ and $m=800$. The mean average
improvement of \MA\ over \GRASPFFR\ is 8.94\% whereas for \MA\ over \GRASPMou\ 
it is 17.36\%. \refFig{fig:real:080} shows the results of these experiments as boxplots for the RPD from the best solutions
of results obtained by each algorithm on each subset.
For instances with $d = 0.85 \cdot m$,
the best results are again obtained by \MA\ followed by \GRASPMou. The average improvement percentage of
\MA\ over \GRASPMou\ ranges from 4.82\% for $n=200$ and $m=600$ to 11.05\% for $n=100$ and $m=600$.
The mean average
improvement of \MA\ over \GRASPMou\ is 7.77\% whereas for \MA\ over \GRASPFFR\ 
it is 59.25\%. The trends are analogous as are those found in \textsc{RandomSet}, confirming the sustained advantage of \MA\ and the relative performance of \GRASPMou\ and \GRASPFFR.

\refTable{friedman:real} shows statistical results for the Aligned Friedman Rank Test (distributed
according to $\chi^2$ with 2 degrees of freedom: 1324.00). The $p$-value calculated by this test
is 0, and this indicates strong evidence for rejecting the null hypothesis that states equality of
rankings between the populations. We can also observe that \MA\ is ranked as the best performing
algorithm followed by \GRASPMou\ and \GRASPFFR. Due to statistical differences between the different
algorithms, we proceeded to perform post-hoc procedures in order
to compare the control algorithm (\MA) with other algorithms (\GRASPMou\ and \GRASPFFR).
In all cases and with all procedures, the null hypothesis
that states equality between the distributions obtained by the control algorithm
and the other compared algorithm was rejected, thus there are statistical
differences between the control algorithm and those remaining. The adjusted $p$-values
for different post-hoc procedures are shown in \refTable{friedman:real:adjusted:p}
and again the results can be considered statistically significant (let us also note \emph{en passant} that significant results are also obtained if the analysis is replicated separately for different values of $d$, $m$ and $n$).

\begin{figure}[!ht]
\begin{center}
\includegraphics[angle=0,scale=0.375]{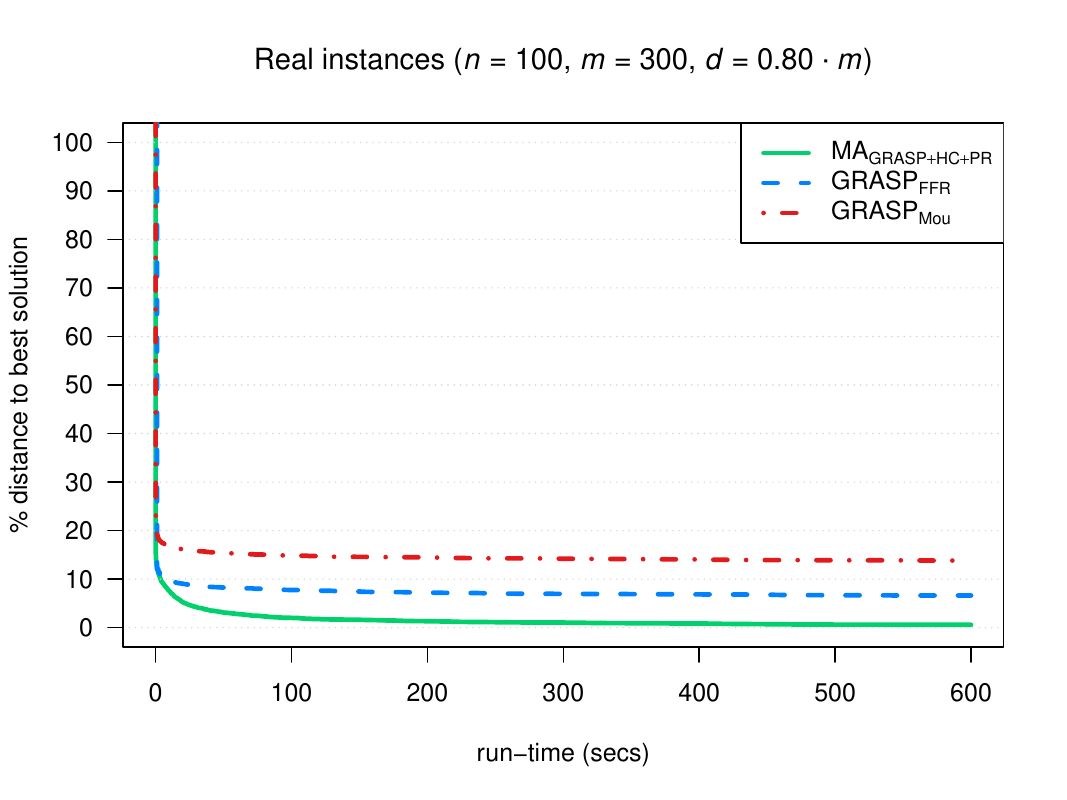}
\includegraphics[angle=0,scale=0.375]{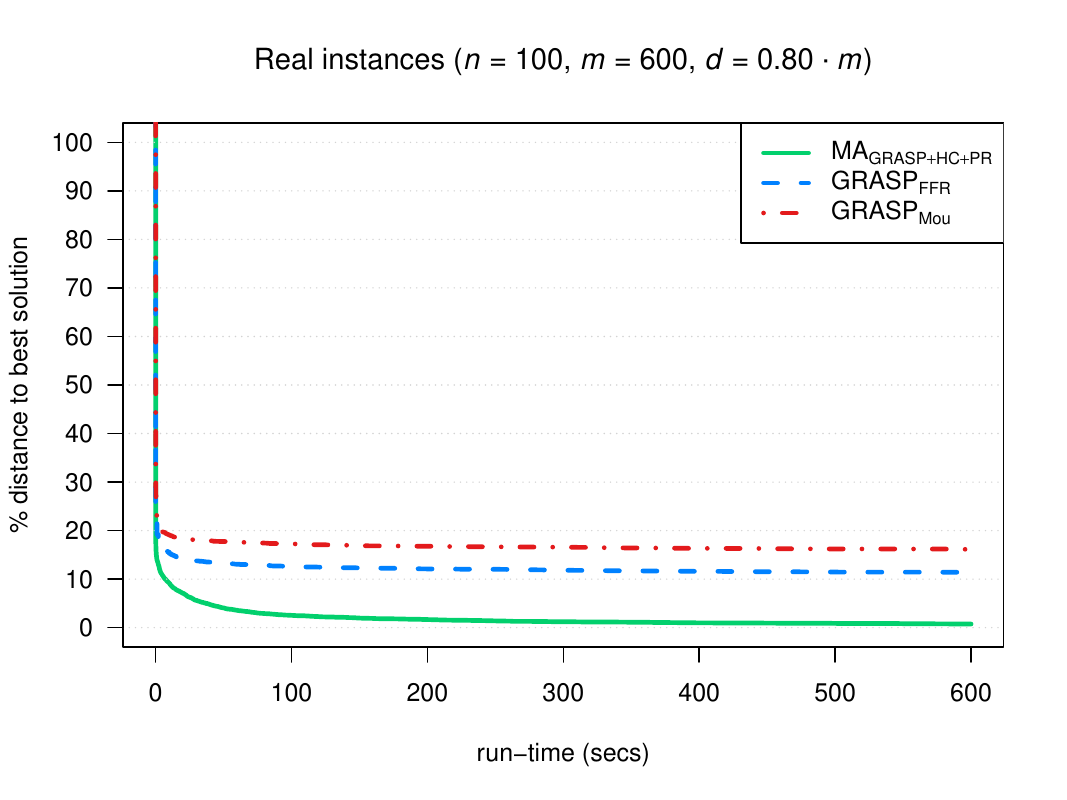}
\includegraphics[angle=0,scale=0.375]{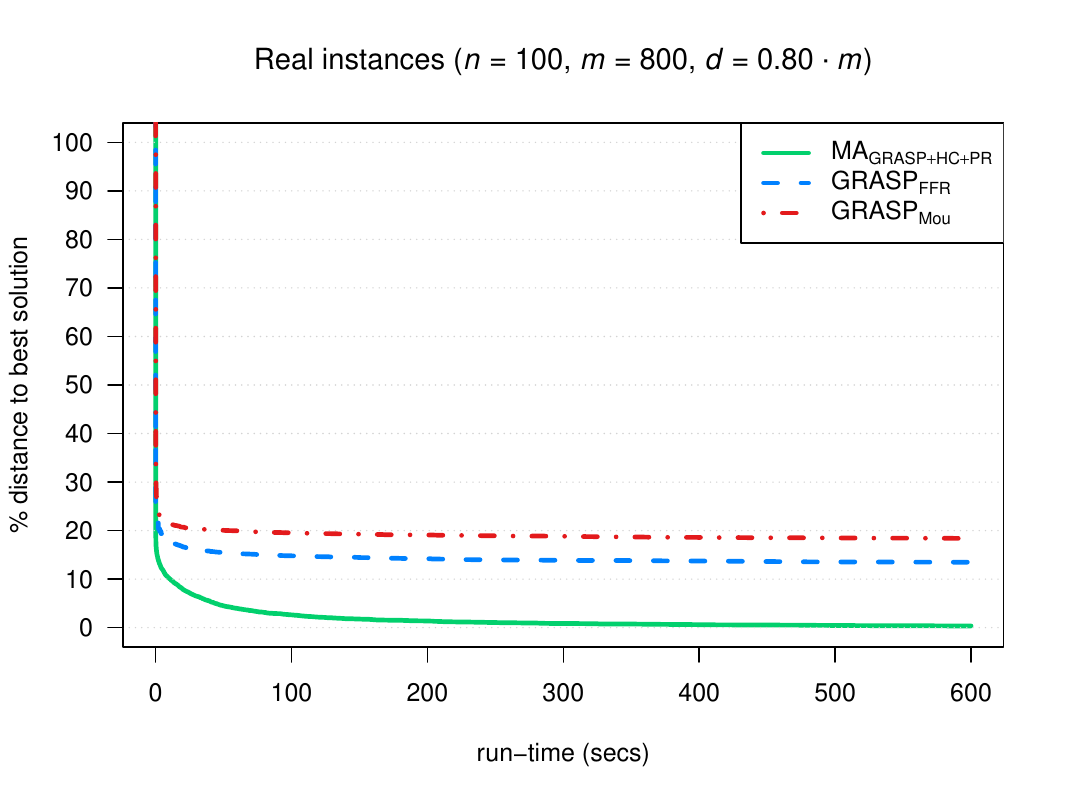}
\includegraphics[angle=0,scale=0.375]{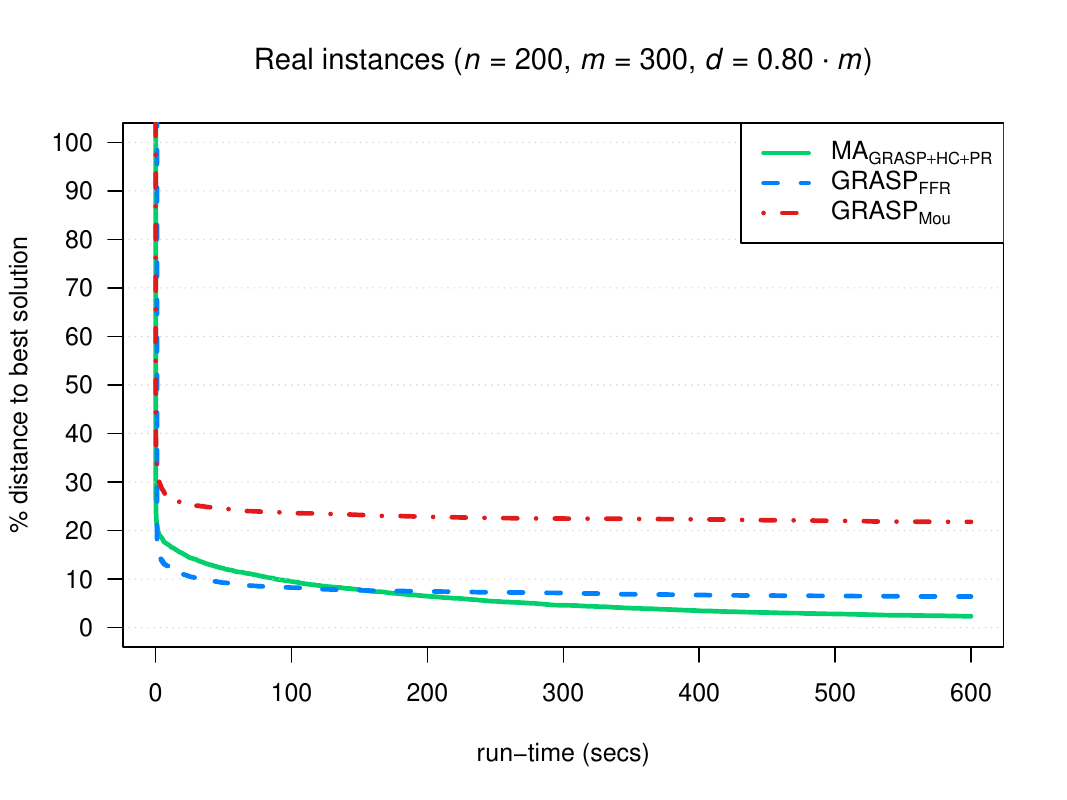}
\includegraphics[angle=0,scale=0.375]{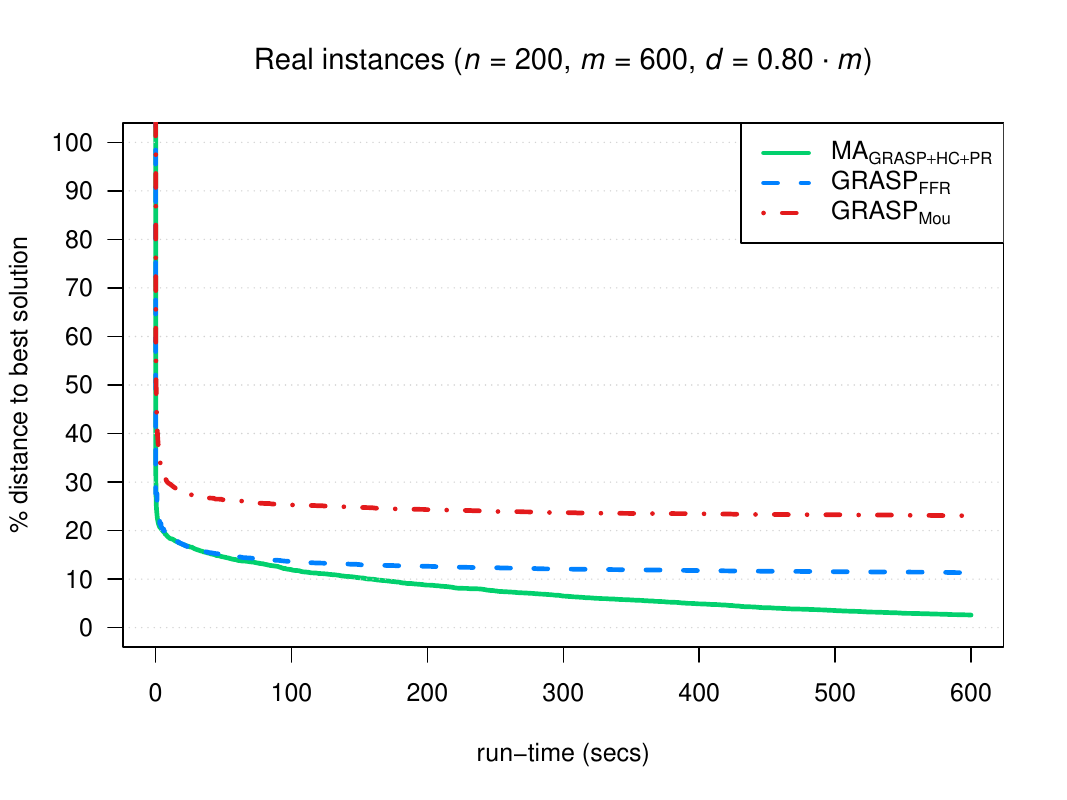}
\includegraphics[angle=0,scale=0.375]{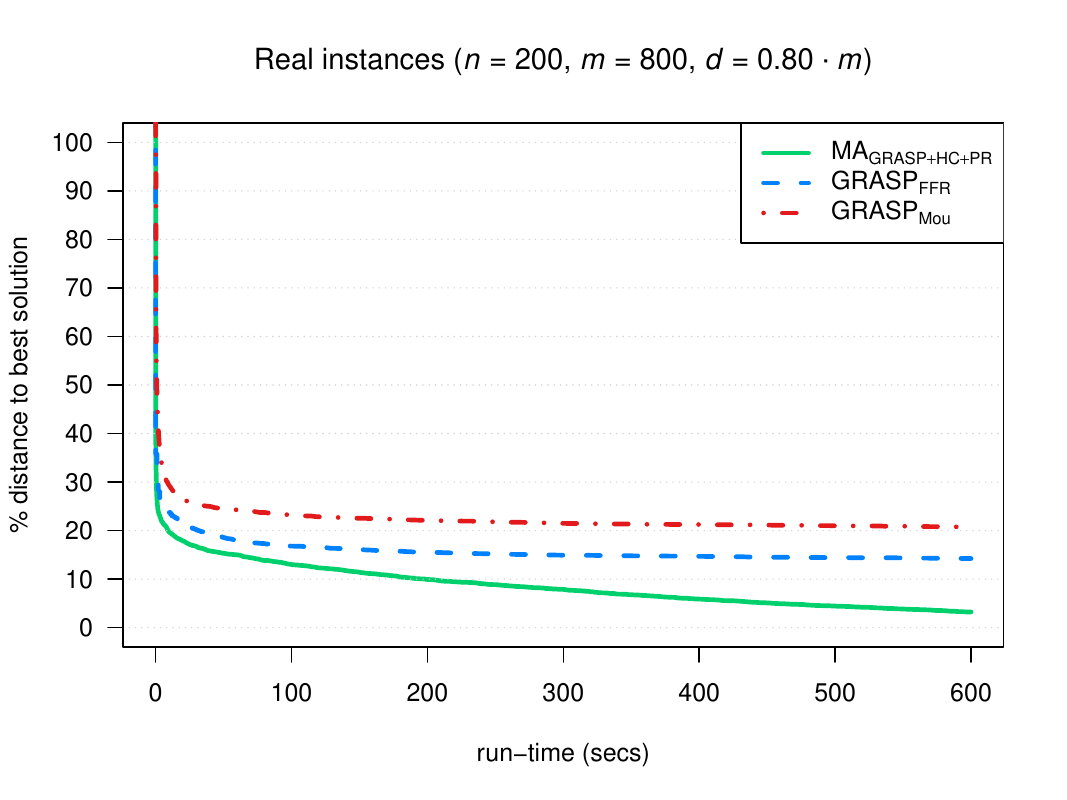}
\end{center}
\caption{Solution quality over time as relative percentage distances (RPD) from best solutions
of results obtained by state-of-the-art algorithms for the FFMSP (\GRASPFFR\ and \GRASPMou) and \MA\
for instances in \textsc{RealSet}
and distance threshold $d=0.80 \cdot m$.
Instances on each dataset are
characterized by a number of strings
($n$) of the same length ($m$) and a distance threshold ($d$).}
\label{fig:real:sqt:080}
\end{figure}

Finally, with the aim of comparing the any-time behavior of the different
algorithms, \refFigs{fig:real:sqt:080}{fig:real:sqt:085} show the solution
quality (as average RPD to best solutions) together with the execution time (SQT)
provided by the different algorithms for instances in \textsc{RealSet} and
the different settings for $n$, $m$ and $d$ (we have omitted SQTs
for $d = 0.75 \cdot m$ as all algorithms reach optimal solutions
right at the beginning of execution). It can be seen that \MA\ produces
better solutions than the remaining algorithms throughout the execution time in all cases,
except for the case of $n=200$, $m=300$ and $d=0.80 \cdot m$, for which
\GRASPFFR\ produces better results for the first 153 seconds of the execution.

\begin{table}[!ht]
\small
\caption{Results obtained by \MA, \GRASPFFR\ and \GRASPMou\ for instances in \textsc{RealSet}. We have considered
18 datasets, each one comprising 5 different instances. Instances on each dataset are characterized by a number of strings
($n$) of the same length ($m$) and a distance threshold ($d$). Table shows mean solution obtained by each algorithm (sol),
along with statistical values (mean ($\mu$) and standard deviation ($\sigma$)) for relative percentage distance (RPD)
from best solutions of results obtained by each algorithm. For \GRASPFFR\ and \GRASPMou\ algorithms, mean improvement
percentage of solutions (imp.\%) obtained by \MA\ with respect to the ones obtained by these algorithms is also shown.}
\tabcolsep=0.06cm
\makebox[\textwidth][c]{
\resizebox{\textwidth}{!}{
\begin{tabular}{rrrrrrrrrrrrrr}
& & & & \multicolumn{2}{c}{\MA} && \multicolumn{3}{c}{\GRASPFFR} && \multicolumn{3}{c}{\GRASPMou}\\
            \cline{5-6} \cline{8-10} \cline{12-14}
$d$ & $n$ & $m$ && sol. & RPD $\mu \pm \sigma$ && sol. & RPD $\mu \pm \sigma$ & imp.\% && sol. & RPD $\mu \pm \sigma$ & imp.\% \\	
\cline{1-3} \cline{5-6} \cline{8-10} \cline{12-14}
$0.75 \cdot m$  & 100 & 300 && 100.00 &   0.00 $\pm$   0.00 && 100.00 &   0.00 $\pm$   0.00 &   0.00 && 100.00 &   0.00 $\pm$   0.00 &   0.00\\
                & 100 & 600 && 100.00 &   0.00 $\pm$   0.00 && 100.00 &   0.00 $\pm$   0.00 &   0.00 && 100.00 &   0.00 $\pm$   0.00 &   0.00\\
                & 100 & 800 && 100.00 &   0.00 $\pm$   0.00 && 100.00 &   0.00 $\pm$   0.00 &   0.00 && 100.00 &   0.00 $\pm$   0.00 &   0.00\\
                & 200 & 300 && 200.00 &   0.00 $\pm$   0.00 && 200.00 &   0.00 $\pm$   0.00 &   0.00 && 200.00 &   0.00 $\pm$   0.00 &   0.00\\
                & 200 & 600 && 200.00 &   0.00 $\pm$   0.00 && 200.00 &   0.00 $\pm$   0.00 &   0.00 && 200.00 &   0.00 $\pm$   0.00 &   0.00\\
                & 200 & 800 && 200.00 &   0.00 $\pm$   0.00 && 200.00 &   0.00 $\pm$   0.00 &   0.00 && 200.00 &   0.00 $\pm$   0.00 &   0.00\\
\cline{1-3} \cline{5-6} \cline{8-10} \cline{12-14}
$0.80 \cdot m$  & 100 & 300 &&  92.66 &   0.57 $\pm$   0.71 &&  87.03 &   6.62 $\pm$   1.57 &   6.04 &&  80.34 &  13.81 $\pm$   1.36 &  13.24\\
                & 100 & 600 &&  96.68 &   0.74 $\pm$   0.57 &&  86.30 &  11.41 $\pm$   1.51 &  10.67 &&  81.70 &  16.14 $\pm$   1.25 &  15.40\\
                & 100 & 800 &&  95.68 &   0.34 $\pm$   0.54 &&  83.05 &  13.48 $\pm$   1.64 &  13.14 &&  78.37 &  18.39 $\pm$   1.46 &  18.05\\
                & 200 & 300 && 126.21 &   2.34 $\pm$   1.58 && 120.92 &   6.40 $\pm$   2.18 &   4.06 && 101.13 &  21.79 $\pm$   2.13 &  19.45\\
                & 200 & 600 && 125.53 &   2.58 $\pm$   1.90 && 114.39 &  11.33 $\pm$   2.53 &   8.74 &&  99.06 &  23.07 $\pm$   1.77 &  20.49\\
                & 200 & 800 && 116.71 &   3.23 $\pm$   1.87 && 103.43 &  14.24 $\pm$   1.91 &  11.01 &&  95.55 &  20.76 $\pm$   1.59 &  17.54\\
\cline{1-3} \cline{5-6} \cline{8-10} \cline{12-14}
$0.85 \cdot m$  & 100 & 300 &&  37.96 &   2.69 $\pm$   2.34 &&  26.45 &  32.30 $\pm$   5.84 &  29.61 &&  34.92 &  10.47 $\pm$   2.03 &   7.78\\
                & 100 & 600 &&  34.23 &   3.81 $\pm$   2.27 &&  10.18 &  71.82 $\pm$   6.83 &  68.01 &&  30.31 &  14.86 $\pm$   1.89 &  11.05\\
                & 100 & 800 &&  31.07 &   4.07 $\pm$   3.12 &&   4.73 &  86.08 $\pm$   7.17 &  82.01 &&  27.61 &  14.65 $\pm$   2.39 &  10.58\\
                & 200 & 300 &&  47.98 &   4.43 $\pm$   2.65 &&  32.66 &  35.39 $\pm$   8.39 &  30.97 &&  44.12 &  12.03 $\pm$   2.45 &   7.60\\
                & 200 & 600 &&  40.65 &   4.52 $\pm$   2.47 &&  16.69 &  63.55 $\pm$  14.51 &  59.02 &&  38.50 &   9.34 $\pm$   2.25 &   4.82\\
                & 200 & 800 &&  31.48 &   4.00 $\pm$   2.56 &&   3.41 &  89.86 $\pm$   4.05 &  85.86 &&  29.91 &   8.82 $\pm$   2.19 &   4.82\\
\cline{1-3} \cline{5-6} \cline{8-10} \cline{12-14}
\end{tabular}
}
}
\label{tab:real}
\end{table}

\begin{table}[!ht]
\small \centering
\caption{Results for the Aligned Friedman Rank Test on \textsc{RealSet} instances
for \MA\ and different state-of-the-art algorithms.}
\begin{tabular}{rlr}
position & algorithm & ranking\\
\hline
1\textsuperscript{st} & \MA       & 1447.82\\
2\textsuperscript{nd} & \GRASPMou & 3014.03\\
3\textsuperscript{rd} & \GRASPFFR & 3639.63\\
\hline
\end{tabular}
\label{friedman:real}
\end{table}

\begin{table}[!ht]
\small \centering
\tabcolsep=0.06cm
\caption{Adjusted $p$-values for $N \times 1$ comparisons of control algorithm (\MA) with
different state-of-the-art algorithms for the Aligned Friedman Rank Test on \textsc{RealSet} instances.}
\resizebox{\textwidth}{!}{
\begin{tabular}{lrrrrrrrr}
algorithm&Bonf&Holm&Hoch&Homm&Holl&Rom&Finn&Li\\
\hline
\GRASPFFR&0.0&0.0&0.0&0.0&0.0&0.0&0.0&0.0\\
\GRASPMou&2.99E-199&1.49E-199&1.49E-199&1.49E-199&0.0&1.49E-199&0.0&1.49E-199\\
\hline
\end{tabular}
}
\label{friedman:real:adjusted:p}
\end{table}

\begin{figure}[!ht]
\begin{center}
\includegraphics[angle=0,scale=0.375]{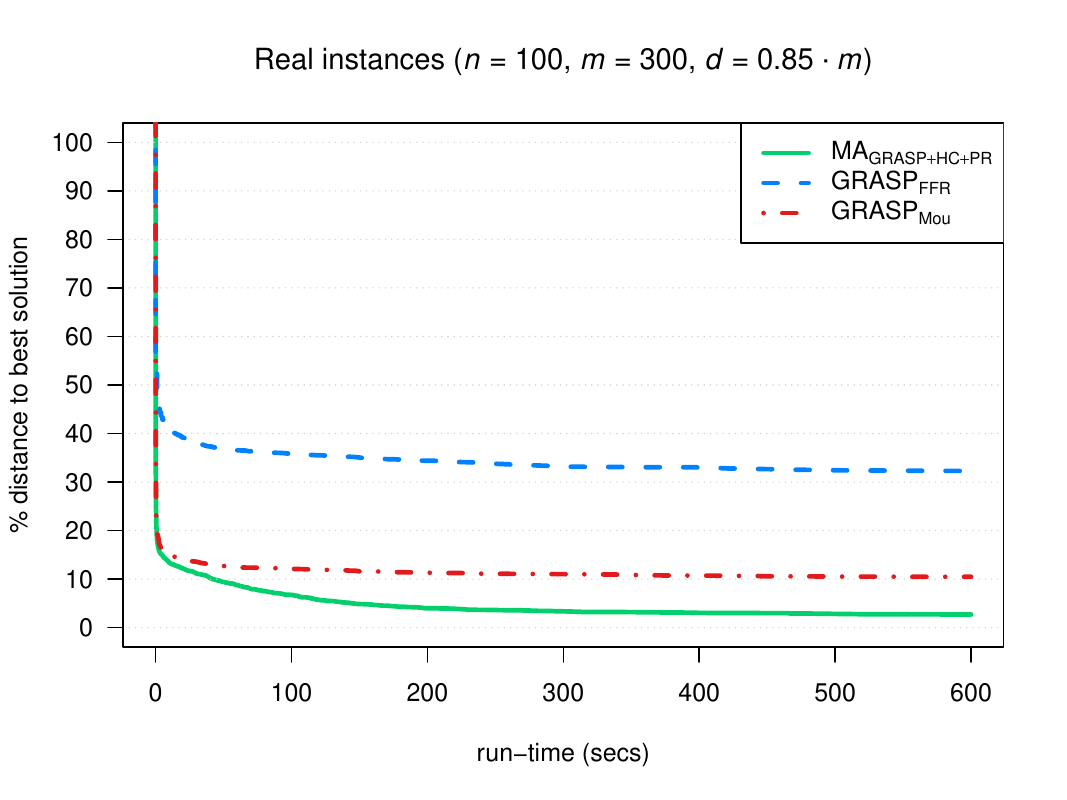}
\includegraphics[angle=0,scale=0.375]{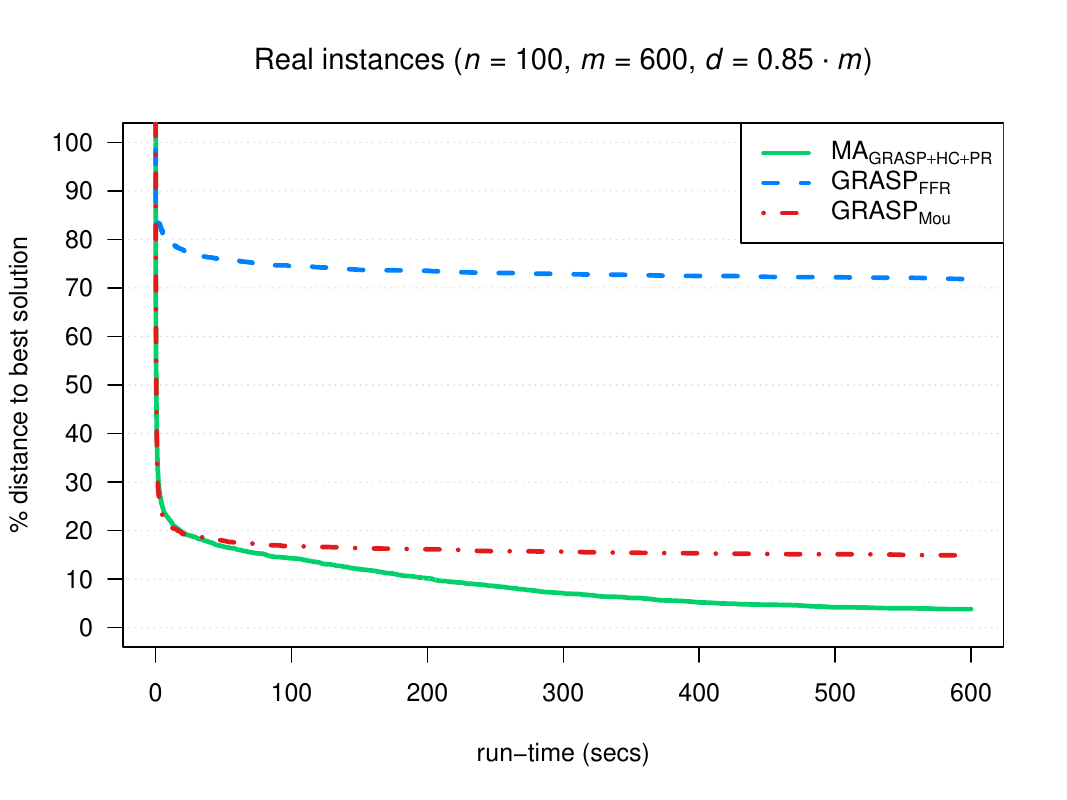}
\includegraphics[angle=0,scale=0.375]{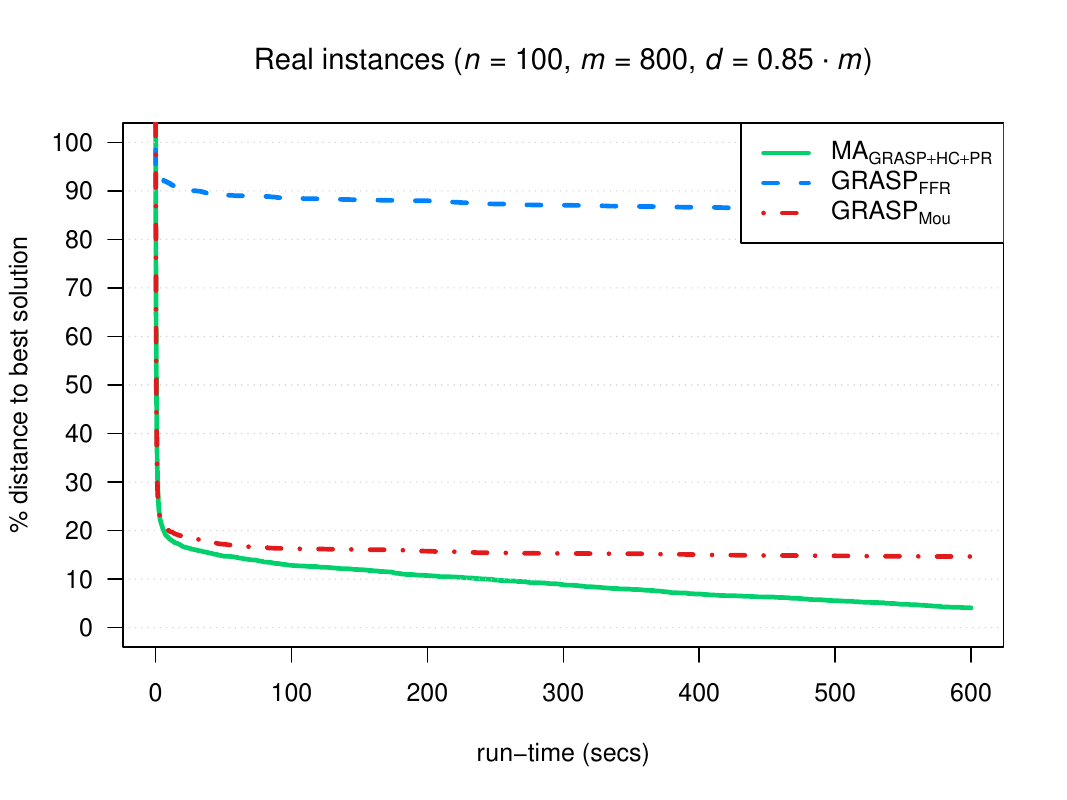}
\includegraphics[angle=0,scale=0.375]{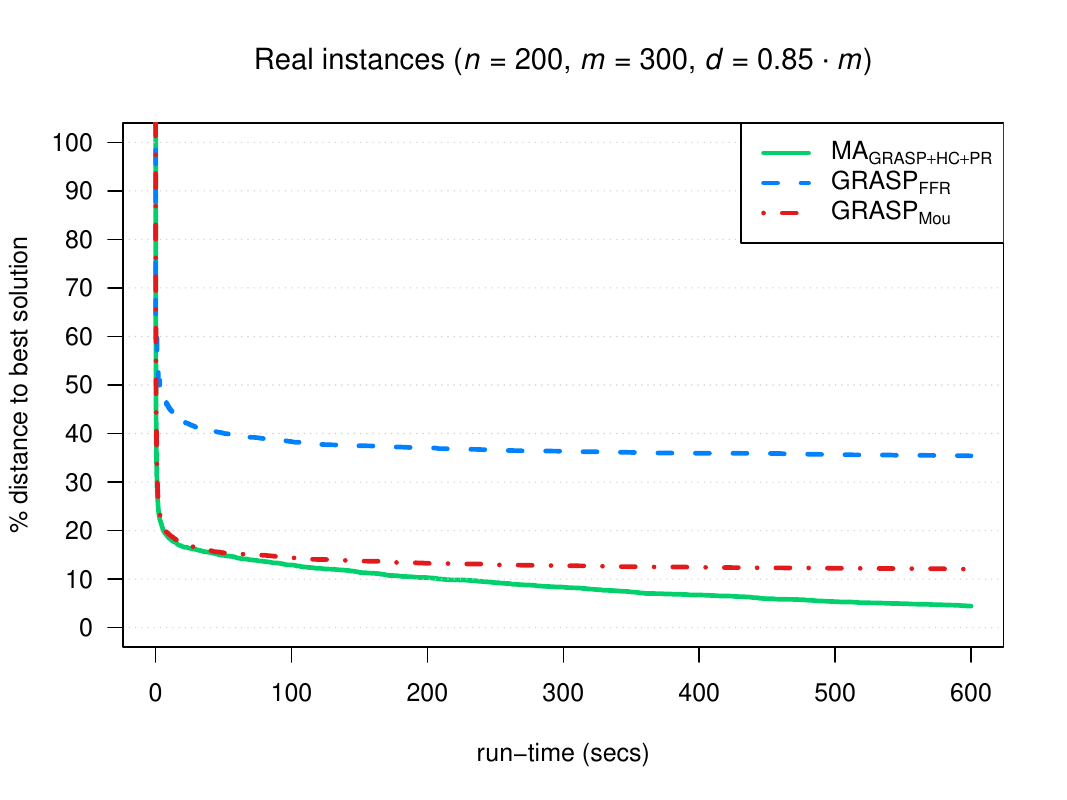}
\includegraphics[angle=0,scale=0.375]{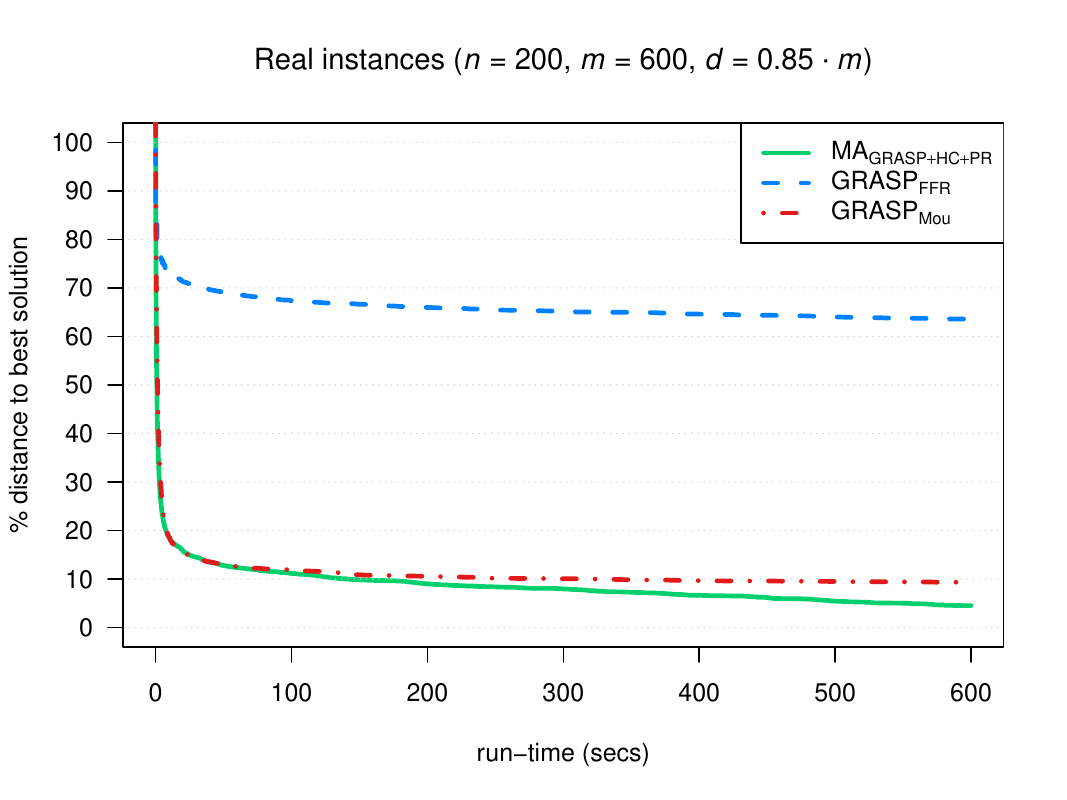}
\includegraphics[angle=0,scale=0.375]{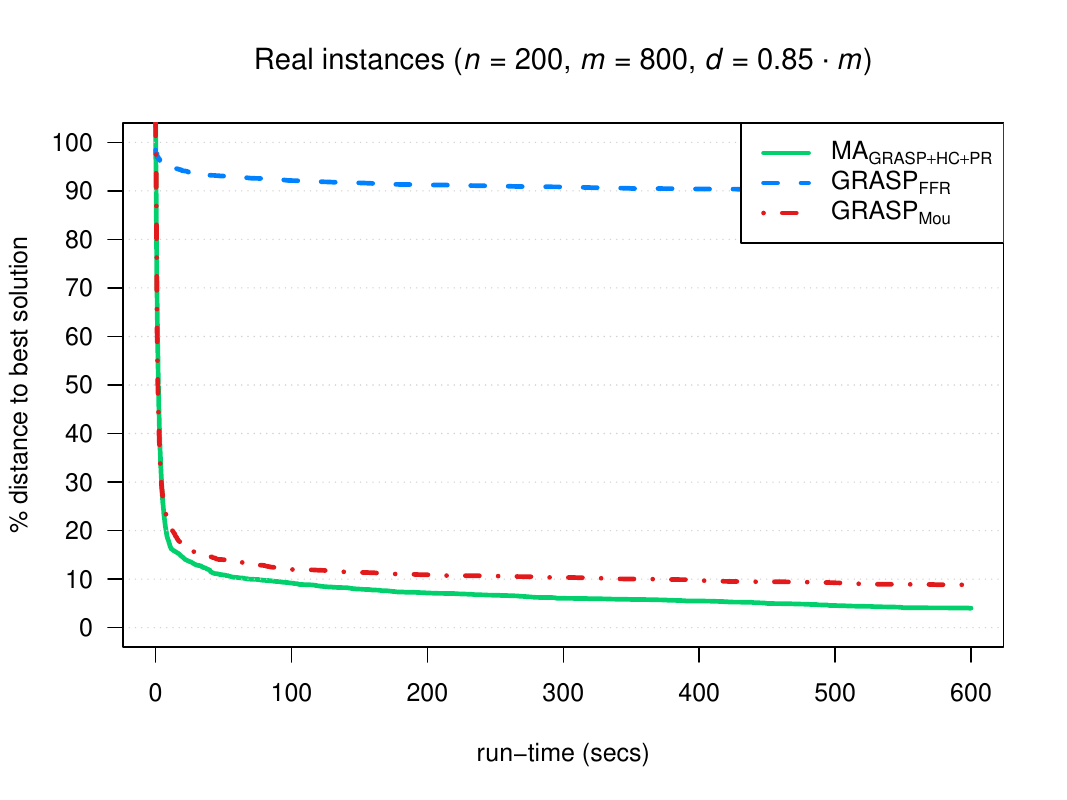}
\end{center}
\caption{Solution quality over time as relative percentage distances (RPD) from best solutions
of results obtained by state-of-the-art algorithms for the FFMSP (\GRASPFFR\ and \GRASPMou) and \MA\
for instances in \textsc{RealSet}
and distance threshold $d=0.85 \cdot m$.
Instances on each dataset are
characterized by a number of strings
($n$) of the same length ($m$) and a distance threshold ($d$).}
\label{fig:real:sqt:085}
\end{figure}
\clearpage

\subsection{Very Large Real Instances}

Finally, we have compared our proposal with state-of-the-art algorithms
on very large real instances. For this benchmark, nine subsets of instances
have been defined using 400 strings ($n$=400) and different
string lengths ($m \in \{1600, 2400, 3200\}$)
and distance thresholds ($d \in \{0.75 \cdot m, 0.80 \cdot m, 0.85 \cdot m\}$). 
For each of the nine subsets, five different
instances have been generated yielding thus a total of 45 instances.

The results of this comparison are shown in \refTable{tab:real:very:large}. It can be observed 
that for $d = 0.75 \cdot m$ optimal solutions were found in all cases by \MA\ and \GRASPMou.
For $d = 0.80 \cdot m$ and $d = 0.85 \cdot m$, the best results are obtained by \MA. The mean average
improvement of \MA\ over \GRASPFFR\ is 18.12\% whereas for \MA\ over \GRASPMou\
it is 13.65\%. For $d = 0.85 \cdot m$, the mean average
improvement of \MA\ over \GRASPFFR\ is 88.20\% whereas for \MA\ over \GRASPMou\ 
it is 2.55\%.  
It is notable that the performance of \GRASPFFR\ substantially degrades in this case for $d=0.85 \cdot m$, that is, the most restricted scenario. The heuristic fitness function used by \MA\ and \GRASPMou\ seems to be able to cope with this situation better.

The results of the Aligned Friedman Rank Test for these instances are  shown in 
\refTable{friedman:real:very:large}  (distributed
according to $\chi^2$ with 2 degrees of freedom: 652.29). The $p$-value calculated by this test
is $p=2.54\cdot10^{-10}$, and this provides evidence for rejecting the null hypothesis that states equality of
rankings between the populations. 
The results of post-hoc procedures comparing
the control algorithm (\MA) with those remaining are shown in 
\refTable{friedman:real:very:large:adjusted:p} as adjusted $p$-values. The null hypothesis
that states equality between the distributions obtained by the control algorithm
and the other compared algorithms was rejected by all procedures for all algorithms
(results are statistically significant).

\begin{table}[!ht]
\small
\caption{Results obtained by \MA, \GRASPFFR\ and \GRASPMou\ for very large real instances. We have considered
9 datasets, each one comprising 5 different instances. Instances on each dataset are characterized by a number of strings
($n$) of the same length ($m$) and a distance threshold ($d$). Table shows mean solution obtained by each algorithm (sol),
along with statistical values (mean ($\mu$) and standard deviation ($\sigma$)) for relative percentage distance (RPD)
from best solutions of results obtained by each algorithm. For \GRASPFFR\ and \GRASPMou\ algorithms, mean improvement
percentage of solutions (imp.\%) obtained by \MA\ with respect to the ones obtained by these algorithms is also shown.}
\tabcolsep=0.06cm
\makebox[\textwidth][c]{
\resizebox{\textwidth}{!}{
\begin{tabular}{rrrrrrrrrrrrrr}
& & & & \multicolumn{2}{c}{\MA} && \multicolumn{3}{c}{\GRASPFFR} && \multicolumn{3}{c}{\GRASPMou}\\
            \cline{5-6} \cline{8-10} \cline{12-14}
$d$ & $n$ & $m$ && sol. & RPD $\mu \pm \sigma$ && sol. & RPD $\mu \pm \sigma$ & imp.\% && sol. & RPD $\mu \pm \sigma$ & imp.\% \\	
\cline{1-3} \cline{5-6} \cline{8-10} \cline{12-14}
$0.75 \cdot m$  & 400 & 1600 && 400.00 &   0.00 $\pm$   0.00 && 399.94 &   0.02 $\pm$   0.06 &   0.02 && 400.00 &   0.00 $\pm$  0.00 &   0.00\\
                & 400 & 2400 && 400.00 &   0.00 $\pm$   0.00 && 399.98 &   0.01 $\pm$   0.03 &   0.01 && 400.00 &   0.00 $\pm$  0.00 &   0.00\\
                & 400 & 3200 && 400.00 &   0.00 $\pm$   0.00 && 399.91 &   0.02 $\pm$   0.07 &   0.02 && 400.00 &   0.00 $\pm$  0.00 &   0.00\\
\cline{1-3} \cline{5-6} \cline{8-10} \cline{12-14}
$0.80 \cdot m$  & 400 & 1600 && 131.70 &   4.55 $\pm$   2.31 && 112.80 &  18.76 $\pm$   6.69 &  14.21 && 115.28 &  16.38 $\pm$  2.74 &   11.83\\
                & 400 & 2400 && 143.12 &   5.34 $\pm$   3.25 && 119.22 &  22.64 $\pm$  11.44 &  17.30 && 118.99 &  20.88 $\pm$  4.48 &   15.54\\
                & 400 & 3200 && 134.94 &   7.58 $\pm$   3.43 && 102.35 &  30.43 $\pm$   6.45 &  22.85 && 114.83 &  21.15 $\pm$  3.82 &   13.58\\
\cline{1-3} \cline{5-6} \cline{8-10} \cline{12-14}
$0.85 \cdot m$  & 400 & 1600 &&  22.42 &   9.01 $\pm$   5.88 &&   0.61 &  97.51 $\pm$   2.44 &  88.50 &&  21.76 &  11.69 $\pm$  5.70 &   2.67\\
                & 400 & 2400 &&  20.67 &  13.21 $\pm$   5.91 &&   0.16 &  99.31 $\pm$   1.59 &  86.10 &&  20.20 &  15.20 $\pm$  5.67 &   1.99\\
                & 400 & 3200 &&  18.18 &   9.97 $\pm$   5.76 &&   0.00 & 100.00 $\pm$   0.00 &  90.03 &&  17.58 &  12.98 $\pm$  5.62 &   2.98\\
\cline{1-3} \cline{5-6} \cline{8-10} \cline{12-14}
\end{tabular}

}
}
\label{tab:real:very:large}
\end{table}

\begin{table}[!ht]
\small \centering
\caption{Results for the Aligned Friedman Rank Test on very large real instances
for \MA\ and different state-of-the-art algorithms.}
\begin{tabular}{rlr}
position & algorithm & ranking\\
\hline
1\textsuperscript{st} & \MA       & 725.28\\
2\textsuperscript{nd} & \GRASPMou & 1307.96\\
3\textsuperscript{rd} & \GRASPFFR & 2018.26\\
\hline
\end{tabular}
\label{friedman:real:very:large}
\end{table}

\begin{table}[!ht]
\small \centering
\tabcolsep=0.06cm
\caption{Adjusted $p$-values for $N \times 1$ comparisons of control algorithm (\MA) with
different state-of-the-art algorithms for the Aligned Friedman Rank Test on very large real instances.}
\resizebox{\textwidth}{!}{
\begin{tabular}{lrrrrrrrr}
algorithm&Bonf&Holm&Hoch&Homm&Holl&Rom&Finn&Li\\
\hline
\GRASPFFR&7.06E-271&7.06E-271&7.06E-271&7.06E-271&0.0&7.06E-271&0.0&7.06E-271\\
\GRASPMou&2.56E-56&1.28E-56&1.28E-56&1.28E-56&0.0&1.28E-56&0.0&1.28E-56\\
\hline
\end{tabular}
}
\label{friedman:real:very:large:adjusted:p}
\end{table}

\section{Conclusions}
\label{sec:conclusions}
The FFMSP is a SSP of enormous difficulty, whose resolution demands the use of powerful heuristics. We have proposed a memetic algorithm to this end. Our MA has successfully integrated different metaheuristic components, namely, GRASP for population initialization, path relinking for recombination and hill climbing for local improvement, into an evolutionary search engine. A careful sensitivity analysis of the MA indicates that from a global perspective this full-fledged memetic approach (i) exhibits a statistically significant superiority to versions in which some of the components are substituted by standard operators, and (ii) provides better results when the population initialization is done with more intensive (i.e. higher greediness) GRASP. The comparison with state-of-the-art algorithms from \cite{Mousavi12} and \cite{FeroneFR13} is also favorable to the MA, which provides average improvements of 8\% and 74\%, respectively on the hardest random problem instances and of 8\% and 60\% on 
the hardest real-world instances. Statistical tests confirm the significance of these improvements.

Future work will be directed towards testing the scalability of the MA. For this purpose, we will consider the use of parallel versions of the technique. We also plan to transfer design ideas to related SSPs, in particular to distinguishing SSPs, leading to the actual application of the technique in practical bioinformatic problems.

\section{Acknowledgements}
Authors wish to thank D. Ferone, Dr. Festa and Dr. Resende for providing us with an implementation of their hybrid GRASP algorithm for the FFMSP and to anonymous reviewers for their suggestions and comments. Support from projects ANYSELF (TIN2011-28627-C04-01 -- \url{http://anyself.wordpress.com}) of MICINN and DNEMESIS (TIC-6083 -- \url{http://dnemesis.lcc.uma.es/wordpress}) of Junta de Andaluc\'{\i}a is acknowledged.

\end{document}